\documentclass{article}

\usepackage[preprint,nonatbib]{neurips_2024}

\usepackage[utf8]{inputenc} 
\usepackage[T1]{fontenc}    
\usepackage{amsmath}
\usepackage{mathtools}
\usepackage{bm}
\usepackage{hyperref}       
\usepackage{url}            
\usepackage{booktabs}       
\usepackage{amsfonts}       
\usepackage{nicefrac}       
\usepackage{microtype}      
\usepackage{xcolor}         
\usepackage{bm}
\usepackage{ulem}

\newcommand{\mean}[1]{\left\langle #1\right\rangle}

\definecolor{lightblue}{rgb}{0.27,0.4,0.87}
\definecolor{darkblue}{rgb}{0.02, 0.02, 0.34}
\definecolor{dollarbill}{rgb}{0.0, 0.32, 0.14}
\definecolor{jasper}{rgb}{0.64, 0.23, 0.24}

\newcommand{\caja}[1]{\left[ {#1} \right] }
\newcommand{\paren}[1]{\left( {#1} \right) }

\newcommand{\bs}{\bm{v}}
\newcommand{\bt}{\bm{h}}
\newcommand{\Nv}{N_\mathrm{v}}
\newcommand{\Nh}{N_\mathrm{h}}

\title{Cascade of phase transitions in the training of Energy-based models}

\author{%
  Dimitrios Bachtis$^{1}$ \\
  \And
  Giulio Biroli$^{1}$ \\
  \And
  Aurélien Decelle$^{2,3}$ \\
  \And
  Beatriz Seoane$^{2,3}$
\\[0.5em]
$^1$Laboratoire de Physique de l'Ecole Normale Supérieure, ENS, \\ Université PSL, CNRS, Sorbonne Université, Université Paris Cité, F-75005 Paris, France.\\
$^2$Departamento de Física Teórica I, Universidad Complutense, 28040 Madrid, Spain.\\
$^3$Université Paris-Saclay, CNRS, INRIA Tau team, LISN,
91190, Gif-sur-Yvette, France.
}

\begin{document}

\maketitle

\begin{abstract}
   In this paper, we investigate the feature encoding process in a prototypical energy-based generative model, the Restricted Boltzmann Machine (RBM). We start with an analytical investigation using simplified architectures and data structures, and end with numerical analysis of real trainings on real datasets. Our study tracks the evolution of the model’s weight matrix through its singular value decomposition, revealing a series of phase transitions associated to a progressive learning of the principal modes of the empirical probability distribution.  The model first learns the center of mass of the modes and then progressively resolve all modes through a cascade of phase transitions.  We first describe this process analytically in a controlled setup that allows us to study analytically the training dynamics. We then validate our theoretical results by training the Binary-Binary RBM on real datasets. By using datasets of increasing dimension, we show that learning indeed leads to sharp phase transitions in the high-dimensional limit. Moreover, we propose and test a mean-field finite-size scaling hypothesis. This shows that the first phase transition is in the same universality class of the one we studied analytically, and which is reminiscent of the mean-field paramagnetic-to-ferromagnetic phase transition. 
   
\end{abstract}

\section{Introduction}


In recent years, we have witnessed impressive improvements of unsupervised models capable of generating more and more convincing artificial samples~\cite{ledig2017photo,du2019implicit,dhariwal2021diffusion}.  Although energy-based models~\cite{lecun2006tutorial} and variational approaches~\cite{kingma2013auto} have been in use for decades, the emergence of generative adversarial networks~\cite{goodfellow2020generative}, followed by diffusion models~\cite{sohl2015deep}, has significantly improved the quality of outputs. Generative models are designed to learn the empirical distribution of datasets in a high-dimensional space, where the dataset is represented as a Dirac-delta pointwise distribution. While different types of difficulties are encounter when training these models, there is a general lack of understanding of how the learning mechanism is driven by the considered dataset. This article explores the dynamics of learning in neural networks, focusing on pattern formation. Understanding how this process shapes the learned probability distribution is complex. Previous studies~\cite{ decelle2018thermodynamics, decelle2017spectral} on the Restricted Boltzmann Machine (RBM)~\cite{salakhutdinov2009deep} showed that the singular vectors of the weight matrix initially evolve to align with the principal directions of the dataset, with similar results in a 3-layer Deep Boltzmann Machine~\cite{ichikawa2022statistical}. Additionally, an analysis using data from the 1D Ising model explained weight formation in an RBM with a single hidden node as a reaction-diffusion process~\cite{harsh2020place}. The main contribution of this work is to demonstrate that the RBM undergoes a series of second-order phase transitions during learning, each corresponding to the acquisition of new data features. This is shown theoretically with a simplified model and on correlated patterns; and confirmed numerically with real datasets, revealing a progressive segmentation of the learned probability distribution into distinct parts and exhibiting second order phase transitions.


\section{Related work}


The learning behavior of neural networks has been explored in various settings. Early work on deep linear neural networks demonstrated that even simple models exhibit complex behaviors during training, such as exponential growth in model parameters~\cite{saxe2013exact,saxe2013learning}.
Using singular value decomposition (SVD) of the weight matrix, researchers revealed a hierarchical learning structure with rapid transitions to lower error solutions. Linear regression dynamics later showed a connection between the SVD of the dataset and the double-descent phenomenon~\cite{advani2020high}. Similar dynamics were found in Gaussian-Gaussian RBMs~\cite{decelle2017spectral}, where learning mechanisms led to rapid transitions for the modes of the model's weight matrix. In this context, the variance of the overall distribution is adjusted to that of the principal direction of the dataset, while the singular vectors of the weight matrix are aligned to that of the dataset.
Unlike linear models, non-linear 
neural-networks, supervised or unsupervised ones, can not exhibit partition of the input's space. Yet, linear model in general can not provide a multimodal partition of the input space, should it be in supervised or unsupervised context, at difference with non-linear ones.  

It was then shown that the most common binary-binary RBMs exhibit very similar patterns at the beginning of learning, transitioning from a paramagnetic to a condensation phase in which the learned distribution splits into a multimodal distribution whose modes are linked to the SVD of the weight matrix~\cite{decelle2018thermodynamics}.
The description of this process motivated the use of RBMs to perform unsupervised hierarchical clustering of data~\cite{decelle2023unsupervised,fernandez2024replica}. The succession of phase transitions had been previously observed in the process of training a Gaussian mixture~\cite{rose1990statistical,miller1996hierarchical,bonnaire2021cascade}, and in the analysis of teacher-student models using statistical mechanics~\cite{barkai1994statistical,lesieur2016phase}. The latter cases are easier to understand analytically due to the simplicity of the Gaussian mixture. Nevertheless, the learned features are somewhat simpler, as they are mainly represented by the means and variances of the individual clusters.
Recently, sequences of phase transitions have been used to explain the mechanism with which diffusion model are hierarchically shaping the mode of the reverse diffusion process~\cite{Biroli_2023,biroli2024dynamical,sclocchi2024phase} and due to a spontaneous broken symmetry~\cite{raya2024spontaneous} after a linear phase converging toward a central fixed-point. The common observation is that the learning of a distribution is, in many cases, obtained by a succession of creation of modes performed through a second order process where the variance in one direction first grow before splitting into two parts, and then the mechanism is repeated. This procedure in particular demonstrate a hierarchical splitting, where the system refined at finer and finer scale of features as it adjust its parameters on a given dataset.

\section{Definition of the model}
An RBM is a Markov random field with pairwise interactions on a bipartite graph consisting of two layers of variables: visible nodes (\(\bm{v}= \{v_i, i=1, \ldots, \Nv\}\)) representing the data, and hidden nodes (\(\bm{h}=\{h_j, j=1, \ldots, \Nh\}\)) representing latent features that create dependencies between visible units. Typically, both visible and hidden nodes are binary (\(\{0, 1\}\)), though they can also be Gaussian~\cite{krizhevsky2009learning} or other real-valued distributions, such as truncated Gaussian hidden units~\cite{nair2010rectified}. For our analytical computations, we use a symmetric representation (\(\{\pm 1\}\)) for both visible and hidden nodes to avoid handling biases. However, in numerical simulations, we revert to the standard (\(\{0, 1\}\)) representation. The energy function is defined as follows:
\begin{equation}
    E[\bs,\bt;\bm{W},\bm{b},\bm{c}] = - \textstyle\sum_{ia} v_i W_{ia} h_a -  \sum_i b_i v_i- \sum_a c_a h_a, \label{eq:Def_RBM}
\end{equation}
with $\bm{W}$ the weight matrix and $\bm{b}$, $\bm{c}$   the visible and hidden biases, respectively. The Boltzmann distribution is then given by $p[\bs,\bt| \bm{W},\bm{b},\bm{c}] = Z^{-1} \exp(-E[\bs,\bt; \bm{W},\bm{b},\bm{c}])$ with $Z=\sum_{\{\bs,\bt\}} e^{-E[\bs,\bt]}$ being the partition function of the system.
RBMs are usually trained using gradient ascent of the log likelihood (LL) function of the training dataset $\mathcal{D}=\{\bm{v}^{(1)},\cdots,\bm{v}^{(M)}\}$,  the LL is then defined as
\begin{eqnarray}
    \mathcal{L}(\bm{W},\bm{b},\bm{c}|\mathcal{D})=M^{-1}\sum_{m=1}^M\ln{ p(\bs=\bm{v}^{(m)}|\bm{W},\bm{b},\bm{c})}= M^{-1} \sum_{m=1}^M\ln{\sum_{\{\bt\}} e^{-E[\bm{v}^{(m)},\bt; \bm{W},\bm{b},\bm{c}]}}-\ln{Z} \nonumber
\end{eqnarray}
The computation of the gradient is straightforward and made two terms: the first accounting for the interaction between the RBM's response and the training set, also called \textit{postive term}, and same for the second, but using the samples drawn by the machine itself, also called \textit{negative term}. The expression of the LL gradient w.r.t. all the parameters is  given by
\begin{gather}
   \textstyle \frac{\partial \mathcal{L}}{\partial w_{ia}} = \langle v_i h_a \rangle_{\mathcal{D}} - \langle v_i h_a \rangle_{\mathcal{H}},\,\,\,\, 
    \frac{\partial \mathcal{L}}{\partial b_{i}} = \langle v_i \rangle_{\mathcal{D}} - \langle v_i \rangle_{\mathcal{H}} \,\,\,\, \text{  and  } \,\,\,\, \frac{\partial \mathcal{L}}{\partial c_{a}} = \langle h_a \rangle_{\mathcal{D}} - \langle  h_a \rangle_{\mathcal{H}}, \label{eqs:grad3}
\end{gather}
where $\langle f(\bs,\bt) \rangle_{\mathcal{D}} = M^{-1}\sum_m \sum_{\{\bt\}} f(\bs^{(m)},\bt) p(\bt|\bs^{(m)})$ denotes an average over the dataset, and $\langle f(\bs,\bt) \rangle_{\mathcal{H}}$, the average over the Boltzmann distribution $p[\bm{v},\bm{h};\bm{W},\bm{a},\bm{c}]$. Most of the challenges in training RBMs stem from the intractable negative term, which has a computational complexity of \(\sim \mathcal{O}(2^{{\rm min}(\Nh,\Nv)})\) and lacks efficient approximations. Typically, Monte Carlo Markov Chain (MCMC) methods are used to estimate this term, but their mixing time is uncontrollable during practical learning, leading to potentially out-of-equilibrium training~\cite{decelle2021equilibrium}.

This work focuses on the initial phase of learning and the emergence of modes in the learned distribution from the gradient dynamics given by Eq.~(\ref{eqs:grad3}). In the following section, we first analytically characterize the early dynamics in a simple setting, showing how it undergoes multiple second-order phase transitions. We then numerically investigate these effects on real datasets.

\section{Theory of learning dynamics for simplified high-dimensional models of data}
We develop the theoretical analysis by focusing on simplified high-dimensional probability distributions that concentrate around different regions, or {\it lumps}, in the space of visible variables. Our aim is to analyze how the RBM learns the positions of these lumps, which represent, in a simplified setting, the features present in the data. In order to simplify the analysis, we will consider the Binary-Gaussian RBM (BG-RBM) defined below, yet the same results can be derived for the Binary-Binary RBM (BB-RBM) as shown in the SI~\ref{SI:BinaryBinaryRBM}.

\subsection{Learning two features through a phase transition} \label{sec:dynamicalGauss}
\begin{figure}
    \centering
    \includegraphics[scale=0.47]{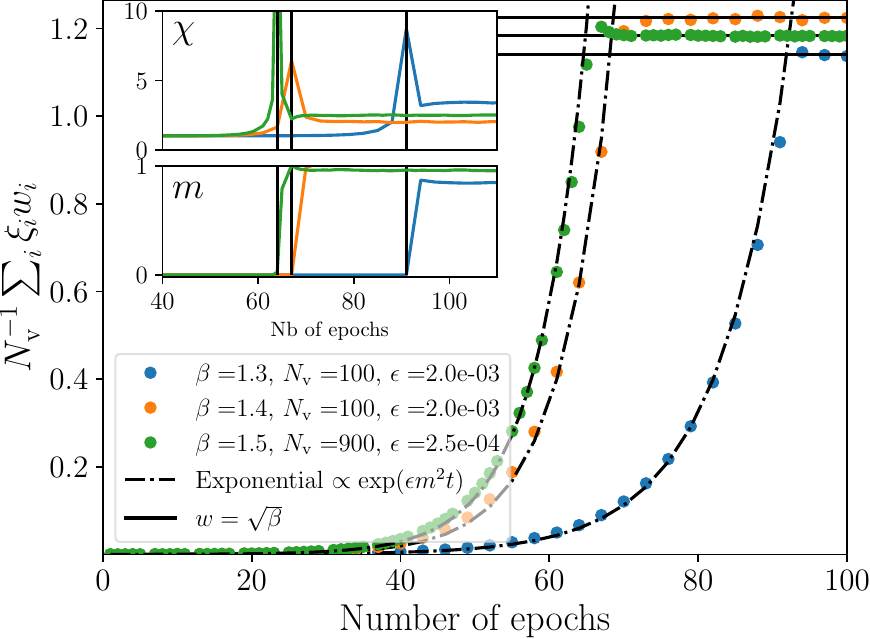}
    \includegraphics[scale=0.47]{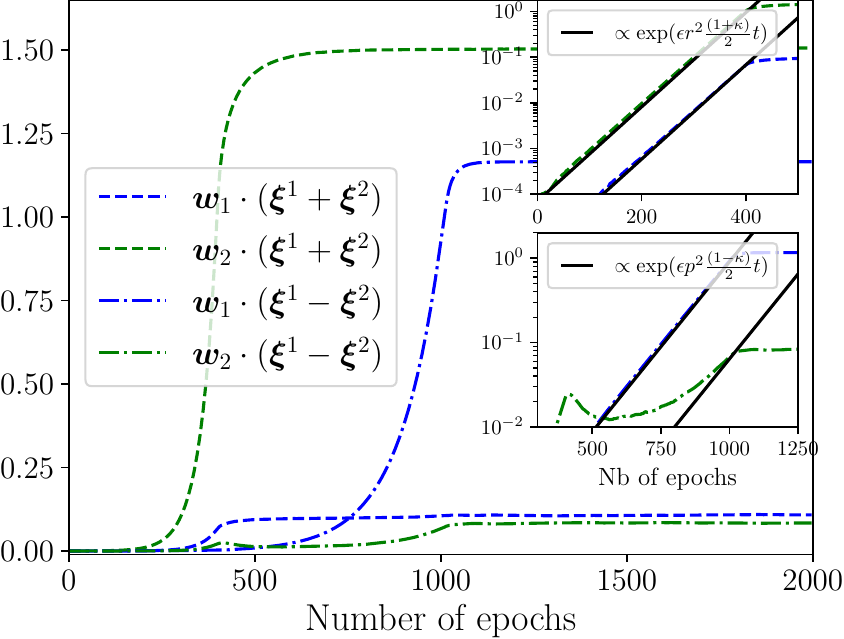}
    \caption{
Learning behavior of the BG-RBM with one hidden node, using data from the Mattis model at different inverse temperatures, system sizes and learning rates \(\beta,\Nv,\epsilon\). The argument of the exponential curves is set to \(m^2 \epsilon \Nv\), where \(\epsilon\) is the learning rate.
\textit{Inset:} (top) behavior of the susceptibility \(\chi\) (bottom) magnetization \(h^*\) of the learning RBM. The vertical line marks the point at which the susceptibility diverges, indicating the onset of spontaneous magnetization. \textbf{Right:} Learning curves for RBMs learning two correlated patterns. The dashed curves represent the weights of the two hidden nodes projected onto \(\bm{\xi}^1+\bm{\xi}^2\), while the dashed-dotted curves are projected onto \(\bm{\xi}^1-\bm{\xi}^2\). \textit{Inset:} Exponential growth during the two phases: top shows growth in the direction \(\bm{\xi}^1+\bm{\xi}^2\) at a rate \(r^2(1+\kappa)/2\), and bottom shows growth in the direction \(\bm{\xi}^1-\bm{\xi}^2\) at a rate \(p^2(1-\kappa)/2\). The arguments of the exponentials are not adjusted.}
    \label{fig:simu1}
\end{figure}

We consider the following simplified setting: we will be using $v_i = \pm 1$ visible nodes, Gaussian hidden nodes and put the biases to zero $\bm{b}=0$ and $\bm{c}=0$. As a model of data, we  consider a Mattis model with a preferred direction $\bm{\xi}$ for the ground state, following the distribution
\[ p_{\rm Mattis} (\bm{v})= \frac{1}{Z_{\rm Mattis}} \exp\left( \frac{\beta}{2\Nv} \left( \sum_{i=1}^{\Nv} \xi_i v_i \right)^2 \right), \]
where $\beta = 1/T$ is the inverse temperature  and $\xi_i=\pm 1 $ represents a pattern encoded in the model as a Mattis state~\cite{mattis1976solvable,hopfield1982neural}. In a Mattis model, $\bm{\xi}$ represents a preferred direction of the model for large values of $\beta$, and in this simple case (with only one pattern) there is no need to specify its distribution as long as its elements are $\pm 1$\footnote{The special case in which all elements $\xi_i=1$ is the well-known Curie-Weiss model in ferromagnetism.}. The Mattis model presents a high-temperature phase with a single mode centred over zero magnetization $m=\Nv^{-1} \sum_i \xi_i \langle v_i \rangle = 0 $ for $\beta < \beta_c$ (where $\langle . \rangle$ is the average w.r.t. the Boltzmann distribution) while in the low-temperature regime, $\beta > \beta_c$, the model exhibits a phase transition between two symmetric modes $m = \pm m_0 (\beta)$ ($\beta_c=1$). Henceforth, we shall focus on the regime $\beta>\beta_c$ where the data distribution is concentrated on two lumps.  From the analytical point of view, we can compute all interesting quantities in the thermodynamic limit $\Nv \to \infty$. In order to keep the computation simple, we will characterize here the dynamics of the system when performing the learning using a BG-RBM~\cite{krizhevsky2009learning} with one single hidden node. In our setting we assume that the distribution of the hidden node is centered in zero (i.e. there is no hidden bias) and that the variance is $\sigma_{\mathrm{h}}^2 = 1/\Nv$ (we discuss the reason for the scaling in SI~\ref{SI:BGRBM}). The distribution is then
\begin{equation*}
p_{\rm BG}(\bm{h}, \bm{v}) = \frac{1}{Z_{\rm BG}} \exp\left( \sum_i v_i h w_i - \frac{h^2 \Nv}{2} \right), \; p_{\rm BG}(\bm{v}) = \frac{1}{Z} \exp\left[ \frac{(\sum_i v_i w_i)^2}{2\Nv} \right].
\end{equation*}
Using this model for the learning, the time evolution of the weights is given by the gradient. With BG-RBM we have that $\langle v_i h \rangle_{\mathcal{H}} = \Nv^{-1} \sum_j w_j \langle v_i v_j\rangle_{\mathcal{H}} $ where the last average is taken over a distribution $p_{\rm BG}(\bm{v})$. We can now easily compute the positive and negative term of the gradient w.r.t. the weight matrix. For the positive term we obtain that $\langle v_i v_j \rangle_{\mathcal{D}} = \xi_i \xi_j m^2$ where $m = \tanh\left(\beta m \right)$. The negative term can also be computed in the thermodynamic limit $\langle v_i v_j \rangle_{\rm RBM} = \tanh(h^* w_i) \tanh(h^* w_j)$ with $ h^* = \frac{1}{N}\sum_k w_k \tanh(h^* w_k)$.
If we take the limit of a very small learning rate, we can convert the parameter update rule using the gradient into a time differential equation for the parameters of the RBM, where $t$ is the learning time: 
\begin{equation} \label{eq:smallw}
    \frac{dw_i}{dt} = \textstyle\epsilon \left[\frac{1}{\Nv} \xi_i  \sum_k \xi_k w_k  m^2 - h^* \tanh(h^* w_i) \right],
\end{equation}
with $\epsilon$ the learning rate~\footnote{In the rest of the derivation, we will remove it since it can be absorb in a redefinition of the time.}. We can analyze two distinct regimes for the dynamics. First, assuming that the weights are small at the beginning of the learning, we get that $h^* = 0$. We can then solve the Eq.~(\ref{eq:smallw}) in this regime obtaining the evolution of the weights toward the direction $\bm{\xi}$ by projecting the differential equation on this preferred direction. Defining $U_{\bm{\xi}} = N^{-1/2}\sum_i \xi_i w_i$, we obtain
\begin{equation*}
  \textstyle  \frac{dU_{\bm{\xi}}}{dt} = m^2 U_{\bm{\xi}} \text{, thus } U_{\bm{\xi}} = \textstyle U_{\bm{\xi}}^0 e^{m^2 t}.
\end{equation*}
This illustrates that the weights are growing in the direction of $\bm{\xi}$ while the projection on any orthogonal direction stays constant. As the weights grow larger, the solution for $h^*$ will depart from zero. Then the correlation between the RBM visible variables starts to grow
\begin{equation*}
    \langle v_i v_j \rangle_{\rm RBM} \approx \frac{1}{Z} \textstyle \int dh h^2 w_i w_j \exp\left( -\frac{\Nv h^2}{2} + \sum_k \frac{h^2 w_k^2}{2}\right) = w_i w_j \frac{1}{\Nv \left(1-\sum_k w_k^2 / \Nv  \right)}, 
\end{equation*}
which means that the susceptibility $\chi = \sum_{i,j} \xi_j \xi_i \langle v_i v_j \rangle_{\rm RBM}$, that is, the response of the system w.r.t. an external perturbation, diverges when $\Nv^{-1}\sum_k w_k^2 \sim 1$, thus exhibiting a {\it second order phase} transition during the learning. Interestingly, $\chi$ diverges as a a power law with a (critical) exponent $\gamma=1$ (where $\Nv^{-1}\sum_k w_k^2$ plays here then the role of the inverse temperature in the standard physical models) thus corresponding to the mean-field universality class~\cite{parisi1988statistical}. Finally, we can study the regime where the weights are not small. In that case, we can first observe that the evolution of the directions orthogonal to $\bm{\xi}$ cancel when the weights $\bm{W}$ align totally with the $\bm{\xi}$ at the end of the training. Finally, taking $w_i = \xi_i w, $ the gradient projected along $\bm{\xi}$ at stationarity imposes
\begin{equation*}
    w m^2 = h^* \tanh(h^* w) \text{ and thus } w=\sqrt{\beta} \text{ and } h^* = \sqrt{\beta m}.
\end{equation*}
We confirm the main results of this section numerically in Fig.~\ref{fig:simu1}, showing they hold accurately even for moderate values of \(\Nv\). The sum of the weights grows exponentially, following the magnetization squared (considering the learning rate), and the weights align with the direction \(\bm{\xi}\), while the norm of the weight vector converges towards \(\sqrt{\beta}\). Additional analysis details and extended computations for the binary-binary RBM case, which is slightly more involved, are provided in the SI.

\subsection{Learning multiple features though a cascade of phase transitions} \label{sec:corr_patt}
We consider now the case in which the data are characterized by more than two features. For concreteness, we focus on the case in which the data is drawn from the probability distribution of the Hopfield model~\cite{hopfield1982neural} with two patterns $\bm{\xi}^1$ and $\bm{\xi}^2$, using the Hamiltonian $\mathcal{H}_H[\bm{v}]= -\frac{\beta}{2\Nv} \sum_{a=1}^2\left( \sum_{i=1}^{\Nv} \xi_i^a v_i \right)^2$. The generalization to a larger (but finite) number of patterns is straightforward. 
Following \cite{tamarit1991pair} we consider the case in which the patterns are correlated and defined as: $\bm{\xi}^1=\bm{\eta}^1+\bm{\eta}^2$ and $\bm{\xi}^2=\bm{\eta}^1-\bm{\eta}^2$; $\bm{\eta}^1$ is a vector whose first $\Nv\frac{1+\kappa}{2}$ components are equal to $\pm1$ with equal probability, and the remaining ones are zero ($0<\kappa<1$). Whereas $\bm{\eta}^2$ is a vector whose last $\Nv \frac{1-\kappa}{2}$ components are equal to $\pm1$ with equal probability, and the remaining ones are zero. When $T<1-\kappa$ this model is in a symmetry broken phase in which the measure is supported by four different lumps centred in $\pm \bm{\xi}^1$ and $\pm \bm{\xi}^2$.  Analogously to what was done previously, we now consider a BG-RBM with a number of hidden nodes equal to the number of patterns where again both hidden nodes are centred in zero and have variance $\sigma_h^2 =1/\Nv$. The Hamiltonian is then given by $\mathcal{H}[\bm{v},\bm{h}] = -\sum_{ia} v_i h_a w_{ia} + \sum_a h_a^2 \Nv/2$, which corresponds to a Hopfield model~\cite{hopfield1982neural} with patterns $\bm{w}^1$ and $\bm{w}^2$. 
The analysis presented in the previous section can be generalized to this case (see SI for more details) and one finds the dynamical equations for the evolution of the patterns:
\begin{align}\label{eq:w}
 \frac{dw_i^a}{dt}=\frac{1}{\Nv} \sum_j\langle v_i v_j \rangle_{\mathcal{D}}w_j^a-\frac{1}{\Nv} \sum_j\langle v_i v_j \rangle_{\rm RBM}w_j^a   
\end{align}
As shown in the SI~\ref{SI:corr_patt}, $\langle v_i v_j \rangle_{\mathcal{D}}=r^2 \eta^1_i \eta^1_j +p^2 \eta^2_i \eta^2_j$ where $r,p$ are a function of $\beta$ (and $\beta^{-1}=T<1-\kappa$, $r>p$). Note that this factorization of the correlation matrix is precisely its spectral eigendecomposition, which means that $\bm{\eta}^1$ and $\bm{\eta}^2$ are nothing but the principal directions of the standard principal component analysis (PCA). At the beginning of the training dynamics the RBM is in its high-temperature disordered phase, hence the second term of the RHS of Eq. (\ref{eq:w}) is zero. The weights $\bm{w}^1$ and $\bm{w}^2$ have therefore an exponential growth in the directions $\bm{\eta}^1$ and $\bm{\eta}^2$, whereas the other components do not evolve. If the initial condition for the weights is very small, as we assume for simplicity, one can then write:
\[\textstyle\bm{w}^a(t)=\frac{z^a}{\sqrt{\Nv\left(\frac{1+\kappa}{2}\right)}}e^{r^2 \left(\frac{1+\kappa}{2}\right)t}\bm{\eta}^1 + \frac{\tilde{z}^a}{\sqrt{\Nv\left(\frac{1-\kappa}{2}\right)}} e^{p^2 \left(\frac{1-\kappa}{2}\right) t}\bm{\eta}^2\qquad a=1,2\,\,\,\,,\] 
where we have neglected the small remaining components; $z^a$ and $\tilde z ^a$ are the projections of the initial condition along the directions $\bm{\eta}^1$ and $\bm{\eta}^2$. 
Since $r>p$, on the timescale $(\log \Nv)/\left(r^2(1+\kappa) \right)$ the component of the $\bm{w}^a s$ along $\bm{\eta}^1$ becomes of order one whereas the one over $\bm{\eta}^2$ is still negligible. In this regime, the RBM is just like the one we consider in the previous section with a single pattern: the system will align with a single pattern that is given in that case by $\bm{\eta}^1\propto \bm{\xi}^1+\bm{\xi}^2$, 
and it has a phase transition at the time $t_I$:
\[\textstyle
\frac{e^{2r^2 \left[\frac{1+\kappa}{2}\right]t_I}}{\Nv}\left((z^1)^2+(\tilde{z}^1)^2\right)=1 \,\,,\]
At $t_I$, the RBM learns that the data can be splitted in two groups centred in $\pm \bm{\eta}^1$, but it does not have yet learned that each one of these two groups consist in two lumps centred in $\bm{\xi}^1$ and $\bm{\xi}^2$ (and respectively $-\bm{\xi}^1$ and $-\bm{\xi}^2$). The training dynamics after $t_I$ can also be analyzed: the components of the weight vectors along $\bm{\eta}^1$ evolve and settle on timescales of order one to a  value which is dependent on the initial condition (see the eq. in the SI). In the meanwhile, the components along $\bm{\eta}^2$ keep growing; at a timescale $(\log \Nv)/(p^2 (1-k))$ (quite larger than $t_I$ in the limit $\Nv \to \infty$) they become of order one. In order to analyze easily this regime, let's consider first the simple case in which the initial condition on the weights is such that $\bm{w}^1(0)\cdot \bm{\eta}^2=-\bm{w}^2(0)\cdot \bm{\eta}^2$ and $\bm{w}^1(0)\cdot \bm{\eta}^1=\bm{w}^2(0)\cdot \bm{\eta}^1$. In this case, one can write $\bm{w}^1=A(t)\bm{\eta}^1+B(t)\bm{\eta}^2$ and $\bm{w}^2=A(t)\bm{\eta}^1-B(t)\bm{\eta}^2$. The corresponding RBM is a Hopfield model with log likelihood: 
\[
\sum_a\frac{(\sum_i v_i w_i^a)^2}{2\Nv} =2A(t)^2\frac{(\sum_i v_i \eta_i^1)^2}{2\Nv} +2 B(t)^2\frac{(\sum_i v_i \eta_i^2)^2}{2\Nv} 
\]
At $t_I$, when $\left(\frac{1+\kappa}{2}\right) A(t_I)^2=1$, one has the first transition in which the RBM measure breaks in two lumps pointing in the direction $\pm \bm{\eta}^1$, as we explained above. In this regime $B(t)$ is still negligible but keeps increasing with an exponential rate. Using the results of \cite{tamarit1991pair}, one finds that when $\frac{1-\kappa}{2}B(t_{II})^2=1$, a second phase transition takes place. This defines a time $t_{II}$ at which the probability measure of the RBM breaks from two lumps to four lumps, each one centred around one of the four directions $\pm \bm{\xi}^1,\pm \bm{\xi}^2$. We have considered a special initial condition, but the phenomenon we found is general. In fact, for any initial condition one can show that the dynamical equations have an instability on the timescale $t_{II}$, which generically induces the second symmetry breaking transition. On Fig.~\ref{fig:simu1}, right panel, we illustrate the exponential growth as described by the theory, toward the two directions. In the SI~\ref{sec:corr_patt}, we show how these phase transitions are in very good agreement with previous work~\cite{decelle2017spectral,decelle2018thermodynamics} and how the phase space is split during training time. At the end of the training, the patterns are given by $\bm{w}^1=\xi^1$ and $\bm{w}^2=\xi^2$ modulo a rotation in the subspace spanned by $\bm{\xi}^{1,2}$, since the likelihood is invariant by rotations in this subspace. In fact, we often found that the patterns are not perfectly aligned because we are not forcing the weights to be binary. This analysis can naturally extend to more than two patterns, typically resulting in a cascade of phase transitions. In this process, the RBM progressively learns the data features, starting from a coarse-grained version (just the center of mass) and gradually refining until all patterns are learned. The analysis done on the BG-RBM can of course be repeated on the BB-RBM (as is done for the case with one mode in the SI~\ref{SI:BinaryBinaryRBM}). The main difference at that level between the two models is that in order to have a retrieval phase, the BB-RBM needs to encode the patterns on an extensive number of hidden nodes (proportional to $\Nv$), while the BG-RBM needs only as many patterns as hidden nodes. Both models can match perfectly the dataset in the limit $\Nv \to \infty$, but we might encounter discrepancies for finite size. However, when dealing with real datasets, by construction, the BG-RBM can not reproduce higher-order correlations and therefore is less interesting than the BB case.

\section{Numerical Analysis}\label{sec:num_ana}
\begin{figure}[t!]
    \centering
    \includegraphics[width=1.0\textwidth,trim=30 50 30 65, clip]{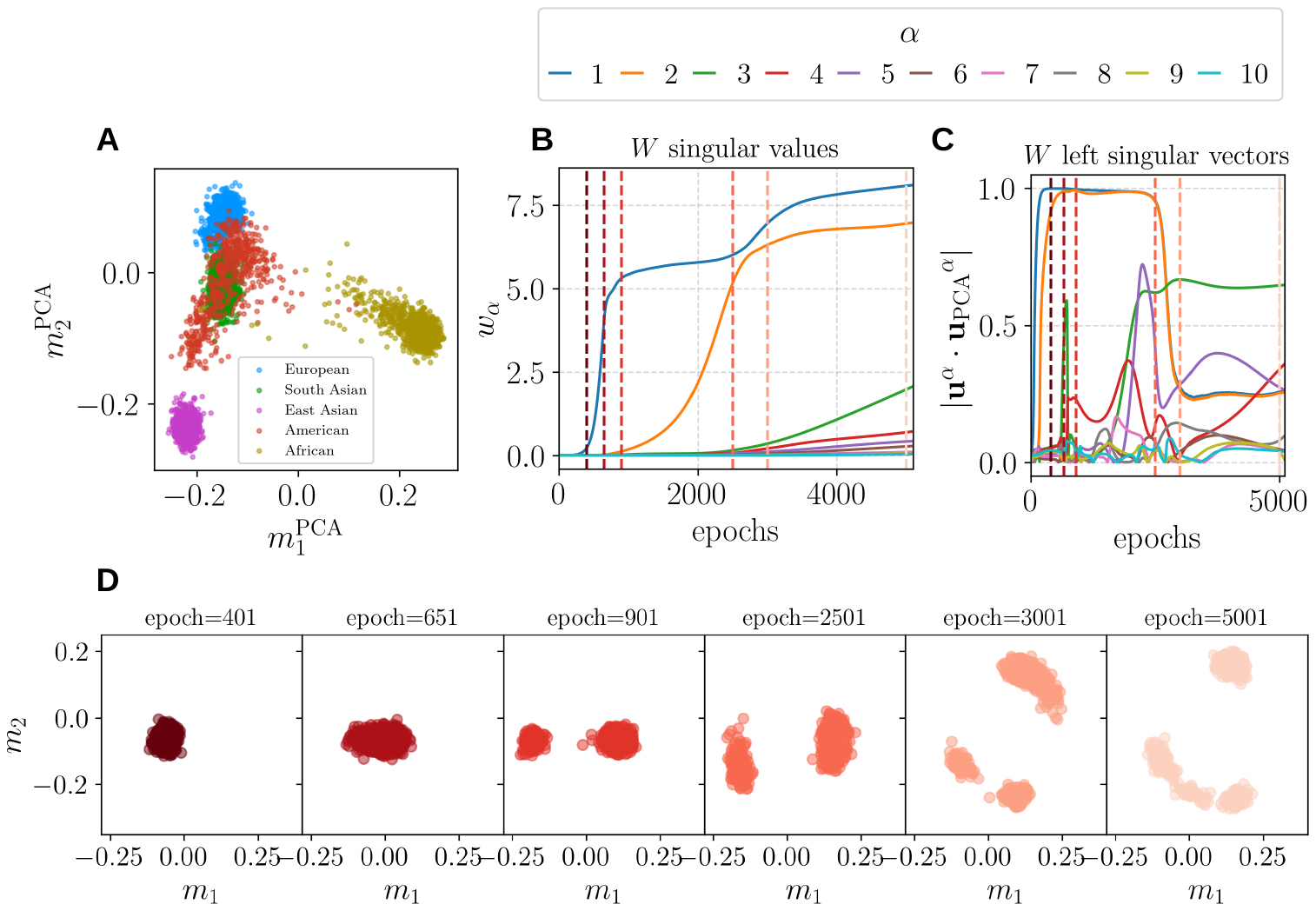}
    \caption{{\bf Human genome dataset.} Progressive coding of the main directions of the dataset when training an RBM with the human genome dataset~\cite{10002015global}. In A, we show the dataset projected along the first two principal components of the dataset $\bm \eta_\alpha$ with $\alpha=1,2$, and $m_\alpha^\text{PCA}=\bm{\eta}_\alpha\cdot \bm x^{(d)}/\sqrt{\Nv}$, with $\bm x^{(d)}$ referring to the different entries in the dataset, i.e. an human individual. Points are colored according to the individual continental origin. In B, we show the evolution of the singular values $w_\alpha$ of the RBM weight matrix $\bm{W}$ as a function of the number of training epochs, and in C, we show the scalar product of the corresponding singular vectors $\bm u_\alpha$ with the corresponding PCA component $\bm \eta_\alpha$.    
    In D, we show the magnetization of the samples generated by the model at different epochs, projected along the first two eigenvectors of $\bm{W}$, which shows that the specialization of the model occurs through the progressive encoding of the main modes of the data in $\bm{W}$. }
    \label{fig:RBM-HGD}
\end{figure}

In the previous sections, we examined the learning process in simplified setups, in order to be able to develop an analytical treatment. In particular, we have shown analytically in a simple setting how the weight matrix is shaped by the patterns present in the dataset and how the learning process dynamics is triggered by the PCA components (the $\bm{\eta}^1$ and $\bm{\eta}^2$ in the previous example) and not by the learning of the encoded patterns ($\bm \xi^1$ and $\bm \xi^2$). Moreover, we have shown that each time the RBMs learn a new direction, the susceptibility of the system diverges with a precise power law everytime the RBM learns a new direction, which is also associated with the development of new modes in the probability measure. In this section we will also show that the insights gained from this simplified analysis are also applicable to understanding the learning process of a Binary-Binary RBM (BB-RBM) with many hidden nodes trained with real data sets. The details about the training procedure are given in SI~\ref{sec:training}.
For this purpose, we will consider 3 real data sets: (i) The Human Genome Dataset (HGD), (ii) MNIST and (iii) CelebA, see details in the SI~\ref{sec:datasets}. To show the occurrence of bonafide phase transitions, it is important to show the effect of increasing the system size (which transforms cross-overs in sharp transitions in the large $\Nv$ limit).  We will therefore resize these data sets in different dimensions by adjusting their resolution, i.e. by changing \(\Nv\) while maintaining comparable statistical properties. Detailed information about the scaling process can be found in the SI~\ref{sec:datasets}.

In real training processes, the machine is expected to gradually learn different patterns \(\bm\xi^\alpha\), from the data, as described in the previous sections. However, the identification of these patterns and their relationship to the statistical properties of the dataset remains unclear. Previous research \cite{decelle2017spectral,decelle2018thermodynamics,decelle2021restricted} has shown that RBM training begins with the stepwise encoding of the most significant principal components of the dataset, \(\{\bm{\eta}^\alpha\}\), which are the eigenvectors of the sample covariance matrix with the highest eigenvalues, on the SVD decomposition of its weight matrix $W_{ia} = \sum_{\alpha} w_\alpha u_i^\alpha \bar{u}_a^\alpha$, where \(\bm{u}^\alpha \in \mathbb{R}^{\Nv}\) and \(\bar{\bm{u}}^\alpha \in \mathbb{R}^{\Nh}\) denote the left and right singular vectors corresponding to the singular value \(w_\alpha\). These vectors form orthonormal bases in \(\mathbb{R}^{\Nv}\) and \(\mathbb{R}^{\Nh}\) respectively, where the index \(\alpha\) ranges from 1 to \(\min(\Nv, \Nh)\) and the singular values \(w_\alpha\) are arranged in descending order. At the beginning of the learning process, the left singular vectors, \(\bm{u}^\alpha\), gradually align $\alpha$-to-$\alpha$ to \(\bm{\eta}^\alpha\). This is consistent with the analytical results in our simple setting in the previous sections. In analogy to the mean-field magnetic models proposed in the previous section, the role of decreasing temperature is played by the increasing magnitude of the singular value \(w_\alpha^2\), associated with each mode $\alpha$, and should lead to a series of phase transitions where the RBM measure splits into increasingly more and more modes. We show in Fig.~\ref{fig:RBM-HGD} that these phenomena are at play by focusing on the evolution of the SVD of the RBM weight matrix when trained with the HGD dataset.

In panel A we show the first two principal components of the dataset, which highlights its strong multimodal structure, as several distant clusters appear (in this case, they are related to the continental origin of the individuals at hand). In Fig.~\ref{fig:RBM-HGD}--B we show the sharp and sudden increases of the singular values $w_\alpha$, as expected from our theoretical analysis, and in Fig.~\ref{fig:RBM-HGD}--C the evolution of the scalar product between $\bm u_\alpha$ and $\bm{\eta}_\alpha$ as a function of the number of training epochs. Different colors indicate different values of $\alpha$. As expected, the modes are progressively expressed during training, and the first two singular vectors match the two principal directions of the dataset for a while. This last figure also shows us that the alignment with the PCA is only temporary (a limitation of current theoretical approaches), as the machine finds better patterns to encode the data as training progresses. 

The progressive splitting of the RBM measure during the training dynamics is shown in Fig.~\ref{fig:RBM-HGD}--D, for which we use $N_\mathrm{s}=1000$ independent samples generated with the model trained up to a different number of epochs (the colors refer to the same epochs highlighted with vertical lines in Figs.~\ref{fig:RBM-HGD}--B and C). For visualization, we show the samples projected onto the right singular vectors of $\bm{W}$, the {\it magnetizations} $m_\alpha=\bm v\cdot \bm u^\alpha/\sqrt{\Nv}$ with $\alpha=1,2$. At the beginning of training, the data points are essentially Gaussian distributed, and the growth of $w_1$ over 4 is related to the splitting of the data into two different clusters on the $m_1$ axis, and the emergence of $w_2$ is related to a second splitting on the $m_2$ axis. At this stage of training, the projections along all subsequent directions are Gaussian distributed as they are the result of a sum of random numbers (fixed by the random initialization of the weight matrix). This progressive splitting is crucial to express the diversity of the dataset shown in Fig.~\ref{fig:RBM-HGD}--A, and can be successfully used to extract relational trees to cluster data points, as recently shown in Ref.~\cite{decelle2023unsupervised}. The details about the numerical analysis are given in~\ref{sec:num-analysis-details}.

At the beginning of training, when only a singular value has been expressed, and thus $\bm W\approx w\bm u \bar {\bm  u}^\top $, the transition of the feature encoding process is analogous to the phase transition from the paramagnetic to the ferromagnetic phase in the Mattis model mentioned above with pattern $\bm{u}$. The detailed justification can be found in SI~\ref{app:RBM_Mattis}. Our analysis allows us to define an effective temperature, linked to the eigenmode of $\bm{W}$ as $\beta = w^2 /16$. Now, since the critical temperature of the Mattis model is $\beta_c=1$, we can show that the BB-RBM will condensate when the first eigenmode of the model reaches $w_c=4$, see SI~\ref{app:RBM_Mattis}.  In a real training, we also have visible $\bm b$ and hidden bias $\bm c$ which could easily change the model towards a random field Mattis model, which leads us to expect a slightly higher critical point but a very similar ferromagnetic phase transition, and in particular, it should not change the transition's mean-field universality class.

\begin{figure}[!t]
    \centering
    \includegraphics[width=1.0\textwidth,trim=30 20 30 20, clip]{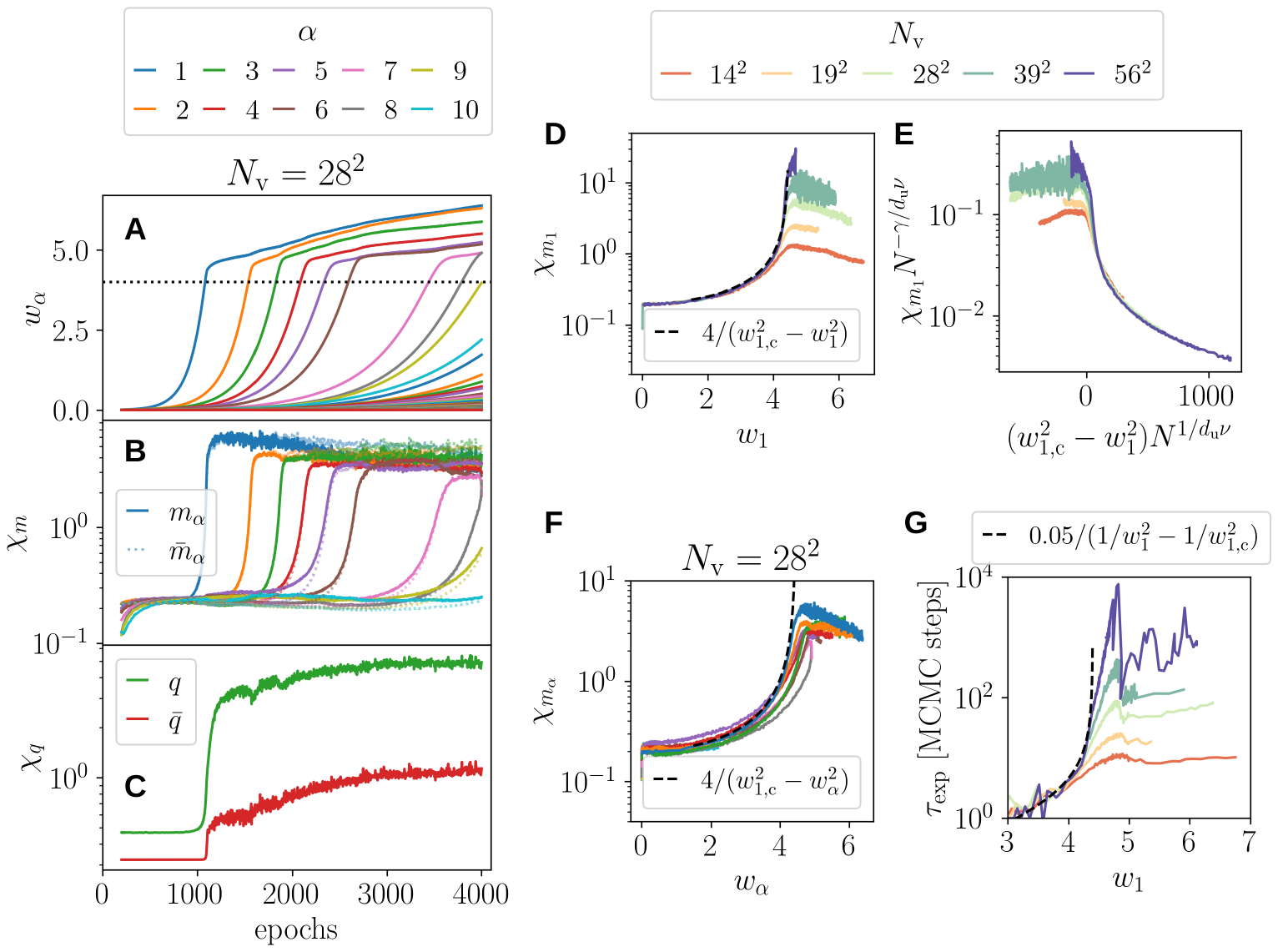}
    \caption{{\bf Traning with the MNIST dataset.} In A we show the evolution of the singular values of the RBM's coupling matrix $\bm{W}$ as a function of the training time. In $B$ we show the evolution of the susceptibilities associated with the magnetizations along the right singular vectors of $\bm{W}$, $m_\alpha=\mean{\bm u_\alpha\cdot \bm v}/\Nv$. In both figures, we consider the standard $\Nv=28^2$ MNIST dataset, different colors refer to different modes. In C we show the susceptibility associated with the overlaps $q$ and $\bar{q}$ between visible and hidden variables. In D we show the susceptibility of the first mode as a function of the first singular value $w_1$ obtained with trainings on MNIST data scaled to different system sizes above and below $L=28$. The numerical curves are compared with the theoretical expectation using the Mattis model in Eq.~\eqref{eq:Mattischi} using $w_{1,\mathrm{c}}=4.45$.
 The same data are shown in E, scaled using the mean-field finite-size scaling ansatz of Eq.~\eqref{eq:FSSchi}. In F, we show the first 10 modes' susceptibilities $\chi_{m_\alpha}$ as a function of their corresponding singular value $w_\alpha$ and compare them with the theoretical curve in D. In G, we show the MCMC relaxation time of the machines trained with different $\Nv$ datasets as a function of $w_1$, together with the theoretical expectation for local moves in dashed lines.
    }
    \label{fig:MNIST}
\end{figure}
To show that there is a cascade of transitions and that what was found for the HGD also holds for other datasets, we now train the RBM with the MNIST dataset. In Fig.~\ref{fig:MNIST}--A we plot the evolution of the singular values $w_\alpha$ along the training, which clearly show the progressive encoding of patterns. The progressive splitting of the RBM measure into clusters and the presence of a phase transition can be monitored by measuring the variance of the distribution of the visible magnetizations $m_\alpha$ along the $\alpha$-th mode or the analogous hidden magnetizations $\bar{m}_\alpha=\bm h\cdot \bar{\bm u}^\alpha/\sqrt{\Nh}$ obtained using the hidden units. The variance of the magnetization multiplied by the number of variables used to compute it and $\beta$, is related to the {\it magnetic susceptibility} via the fluctuation dissipation theorem, which means that
\begin{eqnarray}
    \textstyle\chi_m= \textstyle\Nv\paren{\mean{m^2}-\mean{m}^2} = T\,\mathrm{d}\mean{m}/{\mathrm{d}h} ,
\end{eqnarray}
here $\mean{\cdot}$ refers to the equilibrium measure with respect to RBM's Gibbs measure $p(\bm v, \bm h)$, in practice estimated as the average over $N_\mathrm{s}$ independent MCMC runs. It is well known that the magnetic susceptibility should diverge in the vicinity of a second order phase transition and that such growth in only limited by the overall system size $N=\sqrt{\Nv\Nh}$ in finite systems. These phenomena indeed takes place also in the RBM. We show in 
Fig.~\ref{fig:MNIST}--B 
the evolution of the $\chi_m$s obtained using the magnetizations obtained along the different modes $\alpha$ of $\bm{W}$. As anticipated, the susceptibility $\chi_{m_1}$ associated to the magnetization $m_1$ along the first mode, sharply grows as $w_1$ approaches 4, but it is more remarkable that this behavior is not only restricted to the first mode, but it is also reproduced by the subsequent modes in a step-wise process. According to the mapping between the low-rank RBM and the Mattis (or equivalently, the Curie-Weiss) model, we should expect that our $\chi_m$, at least for the first mode $\alpha=1$ should behave as
\begin{equation}\label{eq:Mattischi}
    \chi_m\sim \frac{4}{\beta_\mathrm{c}-\beta}=\frac{4}{w_\mathrm{c}^2-w^2},
\end{equation}
when approaching the critical point, which is equivalent to stating that the critical exponent is $\gamma=1$. Here, the factor 4 in the numerator is related to the fact that the susceptibility obtained with $\{0,1\}$ variables is 4 times the standard one obtained with Ising spins, and that $\beta=w^2/16$. In Fig.~\ref{fig:MNIST}--D, we show the susceptibility associated with the first mode as a function of $w_1$ using RBMs trained with MNIST data rescaled to different dimensions. As mentioned earlier, the growth of the susceptibility is limited by the system size $\Nv$. However, if we look at increasingly larger sizes, we can observe the growth over several decades. This shows that at the transition we observe the Mattis/Curie-Weiss behavior of Eq.~\ref{eq:Mattischi}, as shown in the black dashed line, where the only adjustable parameter was the critical point $w_\mathrm{1,c}=4.45$ (i.e. there is no adjustable pre-factor).
\begin{figure}[t!]
    \centering
    \includegraphics[width=1.0\textwidth,trim=10 180 30 00, clip]{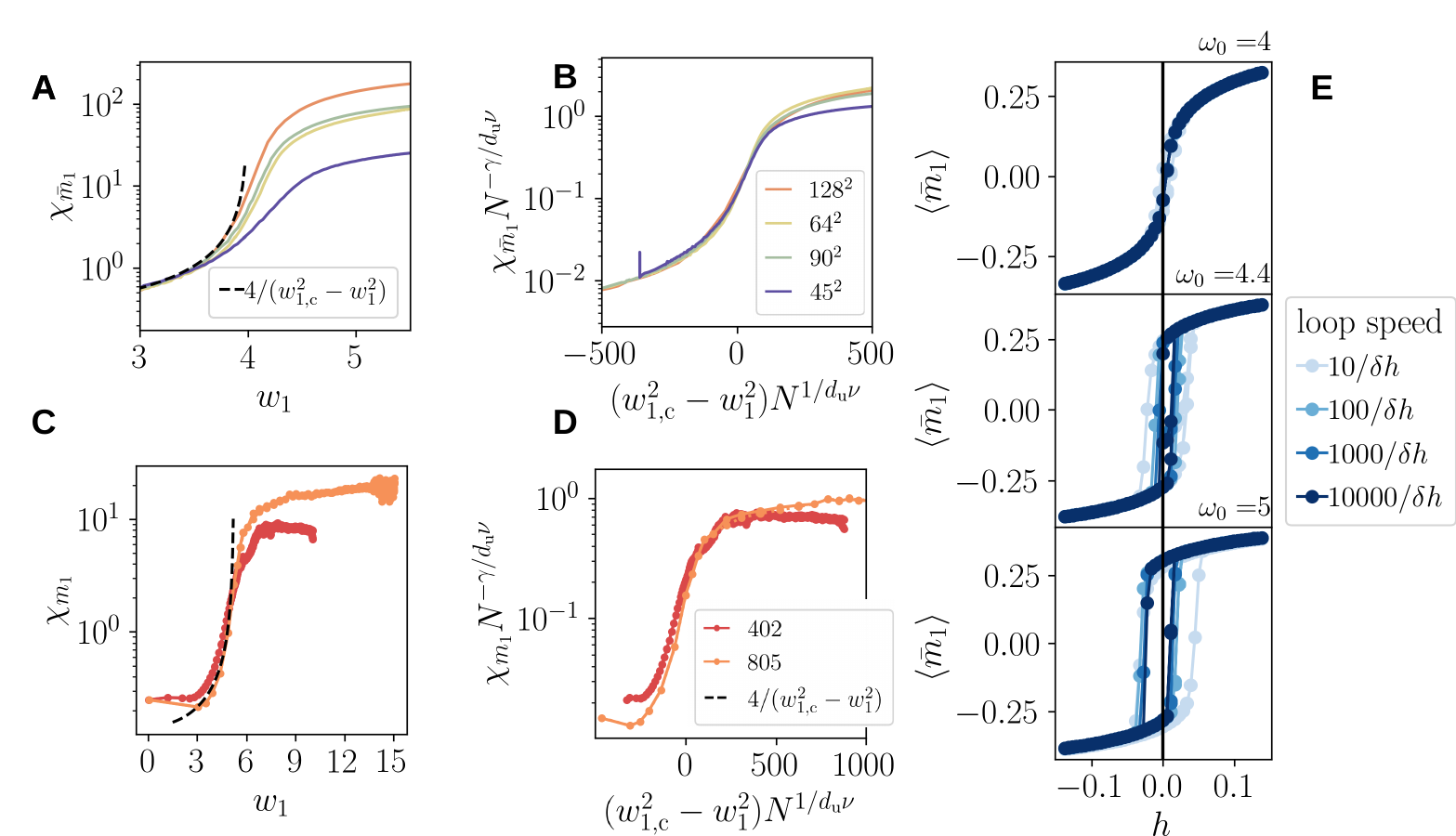}
    \caption{{\bf Training with the CELEBA and HGD datasets:} 
In A, we plot the hidden susceptibility for different system sizes in the CELEBA dataset, with dashed lines indicating the expected divergence at $w_{1,c} = 4$. In B, we show the mean-field FFS associated with the first transition using mean-field exponents. In C and D, we present the visible susceptibility for the first phase transition in the HGD dataset, using $w_{1,c} = 5.25$ for scaling. In E, typical hysteresis in the low-temperature phase is illustrated for CELEBA (128$\times$128), similar to the mean-field Ising model in external fields.}
    \label{fig:CELEBA}
\end{figure}

One of the crucial tests to ensure that a finite-size transition is a bona fide phase transition is to study its behavior by changing the number of degrees of freedom. One of the standard tools to do this is to make use of the so-called finite-size scaling (FSS) ansatz, motivated by renormalization group arguments~\cite{binder1981finite,amit2005field, cardy2012finite}. Mean-field models follow a modified FSS ansatz which was first studied in~\cite{brezin1982investigation}. In particular, the FSS ansatz for the susceptibility is
\begin{equation}\label{eq:FSSchi}
    \textstyle\chi^N_m (\beta)=N^{\frac{\gamma}{\nu d_\mathrm{u}}}\phi\paren{N^{\frac{1}{\nu d_u}} |\beta-\beta_\mathrm{c}|},
\end{equation}
with $\phi(\cdot)$ a size-independent scaling function,  $N=\sqrt{\Nv\Nh}$ is the effective size of our model and $\gamma=1$, $\nu=1/2$ and $d_u=4$ as expected in the mean-field universality class. We test this ansatz in Fig.~\ref{fig:MNIST}--E showing that it does succeed to scale the finite-size data in the critical region, especially in the largest system sizes, which confirms both the mean-field universality class and the prevalence of the transition in the thermodynamic limit. In Fig.~\ref{fig:CELEBA}-A and B, and Fig.~\ref{fig:CELEBA}-C and D, we show that the indicators of a phase transitions--growth of the susceptibility and its mean-field finite size scaling--also holds for CelebA and HGD datasets.  Finally, a final piece of evidence of the existence of a phase transition is presented in Fig.~\ref{fig:CELEBA}-E, where we show that after the continuous transition has taken place, one can induce a discontinuous transition and hysteresis effects by applying a field in the direction of the learned pattern, in full agreement with what observed for standard phase transitions, the details of the analysis can be found in SI~\ref{app:histnum}, and further theoretical insights into the relationship between hysteresis and discontinuous transitions can be found in~\ref{SI:hyst}.

All the discussion so far has been mainly concerned with the first phase transition, when the RBM learns the first mode. But we have discussed in Fig.~\ref{fig:MNIST}--B that an entire sequence of step-wise phase transitions occurred in the rest of the $\bm{W}$-matrix modes. In Fig.~\ref{fig:MNIST}--F we show each of these mode susceptibilities $\chi_{m_\alpha}$ as a function of their corresponding singular value. They show extremely similar divergent behavior with respect to the mode $\alpha=1$, with an apparent slight variation of the critical point for each mode, although all appear to remain close to the predicted $w_\alpha\sim 4$, suggesting that the subsequent transitions may be of similar mean-field nature. Exceeding second-order phase transitions has a very strong impact on the overall quality of the training, in particular on the quality of the log-likelihood gradient estimated by MCMC dynamics. Indeed, second-order transitions are associated with a well-known arresting effect, known as {\it critical slowing-down} behavior, by which the thermalization times diverge with the correlation length $\xi^z\propto |\beta-\beta_\mathrm{c}|^{-\nu z}$, where $z$ is the dynamical critical exponent, which is 2 for local and non-conserved order parameter moves in mean-field, making the thermalization of large systems extremely difficult in practice. We show that our exponential relaxation times diverge exactly as predicted in Fig.~\ref{fig:MNIST}--G. This has a significant impact on the quality of models trained with maximum likelihood approaches, as these methods rely on MCMC to estimate gradients. It is therefore expected that MCMC mixing times increase sharply each time a mode is coded, which can be prohibitive for clustered and high-dimensional datasets. Recent studies have shown that pre-training a low-rank RBM using other methods (and thus bypassing the initial phase transitions) can be very effective in improving the models in clustered datasets~\cite{bereux2024fast}. However, we emphasize that the cascade of phase transitions described in this paper occurs regardless of the training scheme or whether the Markov chains reach equilibrium. This is discussed further in SI~\ref{app:noneq}.

{\bf Extensions and limitations-- }
All these results can be studied in detail for RBMs thanks to the fact that we can analytically deal with the Hamiltonian. However, our results can be extended to the case of Deep Boltzmann Machines, where previous works have also computed the phase diagram, which are also based on the SVD decomposition of the weighing matrices~\cite{ichikawa2022statistical}, but also in diffusion models where phase transitions linked to the learning has also been described~\cite{biroli2024dynamical,raya2024spontaneous}. It therefore stands to reason that similar phenomena occur with even more complex models such as Convolutional EBM, but where it is not clear how the parameters of the model can be decomposed. A first test would be to see what the projection of the generated data would look like in the different phases of learning.

\section{Conclusions}
In this paper, we first characterized the learning mechanism of RBMs using a simplified setting with a dataset provided by a simple teacher model. We used two examples: one with two symmetric clusters and another with four correlated clusters. Our results show that the learning dynamics identify modes by exponential growth in the directions of the clusters dominated by the variances of these clusters. The theory predicts the timing of the first phase transitions and agrees well with~\cite{decelle2018thermodynamics}. Numerically, we have confirmed the existence of a cascade of phase transitions associated with the growing modes \(w_\alpha\) and accompanied by divergent susceptibility. Finite-size scaling suggests that these transitions are critical and fall into the class of mean-field universality. This set of phase transitions likely goes beyond RBMs and offers insights into learning mechanisms, particularly for generative models. These transitions have significant implications for both training and understanding the learned features. During training, each transition is associated with a divergent MCMC relaxation time, which requires careful handling to properly train the model. In addition, the hysteresis phenomenon ensures that the learning trajectory involves second-order phase transitions, which are beneficial for tracking the emergence of modes in the learned distribution. However, changing parameters (such as the local bias) could lead to first-order transitions that are detrimental to sampling and could explain the ineffectiveness of parallel tempering in the presence of temperature changes. In practice, our analysis shows that the principal directions of the weight matrix contain valuable information for understanding the learned model.

\section{acknowledgments}
The authors would like to thank J. Moreno-Gordo for his work and discussions during the initial phase of this project. The authors (A.D. and B.S.) acknowledge financial support by the Comunidad de
Madrid and the Complutense University of Madrid (UCM) through the Atracción de Talento program (Refs. 2019-T1/TIC-13298 and Refs. 2023-5A/TIC-28934), and to project PID2021-125506NA-I00 financed by 
the Ministerio
de Econom\'{\i}a y Competitividad, Agencia Estatal de Investigaci\'on (MICIU/AEI
/10.13039/501100011033), and the Fondo Europeo de Desarrollo Regional (FEDER, UE).

\bibliographystyle{unsrt}
\bibliography{ref}

\appendix

\clearpage

\section{Binary-Gauss RBM}\label{SI:BGRBM}

We add some technical details to the derivation of the dynamical process. We first recall the definition of general BG-RBM first. BG-RBMs consist of a bipartite model where the visible nodes $\bm{v}$ are binary (in our case $\pm 1$) and the hidden nodes are Gaussians. The Hamiltonian follows the same expression as the BB-RBM, Eq.~\ref{eq:Def_RBM}. The hidden nodes are Gaussian random variables centred in zero with variance $\sigma_h^2$. In our analysis where we use only one or few hidden nodes, it is important that the variance scales as the inverse of the system's size $\sigma_h^2 = 1/\Nv$ in order for the Hamiltonian of the system to be an extensive property (proportional to $\Nv$). This scaling is crucial for the analysis: if the energy term scales as the system's size, then it is possible to have two distinct phases: for small weight magnitude, the system is disordered (it does not polarize in any particular direction) while for large values of $\bm{w}$ the system will polarize toward one of the direction encoded in the weight matrix. A more detailed description can be found in~\cite{mattis1976solvable,mezard2017mean}.

Now, recall that we consider the Mattis model, biased toward a pattern $\bm{\xi}$ for generating the dataset
\[ p_{\rm Mattis} (\bm{v})= \frac{1}{Z_{\rm Mattis}} \exp\left( \frac{\beta}{2\Nv} \left( \sum_{i=1}^{\Nv} \xi_i v_i \right)^2 \right) \]
where $\beta$ is the inverse temperature $\beta = 1/T$ and $\xi_i=\pm 1 $ represents a potential pattern direction. The Mattis model presents a high-temperature phase with a single model centred over zero magnetization $m=N^{-1}_{\Nv} \sum_i v_i = 0 $ for $\beta < \beta_c$ while in the low-temperature regime, $\beta > \beta_c$, the model exhibits a phase transition between two symmetric modes $m = \pm m_0 (\beta)$. From the analytical point of view, we can compute all interesting quantities in the thermodynamics limit $\Nv \to \infty$. The RBM's distribution is given by
\begin{align*}
p_{RBM}(\bm{h}, \bm{v}) &= \frac{1}{Z_{RBM}} \exp\left( \sum_i v_i h w_i - \frac{h^2 \Nv}{2} \right) \\
p_{RBM}(\bm{v}) &= \frac{1}{Z} \exp\left( \frac{(\sum_i v_i w_i)^2}{2\Nv} \right)
\end{align*}
Using this model for the learning, the time evolution of the weights is given by the gradient. With BG-RBM we have that
\begin{align}
    \langle v_i h \rangle &= \sum_{\{\bm{v}\}} \int dh v_i h p(\bm{v},h) =  \sum_{\{\bm{v}\}} \int dh v_i h p(h|\bm{v})p(\bm{v}) \\
    &= \frac{1}{\Nv}\sum_{\{\bm{v}\}} v_i \sum_j v_j  p(\bm{v}) w_j =  \frac{1}{\Nv}\sum_j w_j \langle v_i v_j\rangle_p
\end{align}
where the last average is taken over a distribution $p(\bm{v})$. We can now easily compute the positive and negative term of the gradient w.r.t. the weight matrix. For the positive term, assuming that $\beta > 1$, we obtain that
\begin{align*}
  \langle v_i v_j \rangle_{\mathcal{D}} &= \frac{1}{Z_{\rm Mattis}}\int dm \sum_{\{\bm{v}\}} v_i v_j \exp\left( -\beta \Nv\frac{m^2}{2}+m\beta \sum_k \xi_k v_k\right) \\
  &= \frac{1}{Z_{\rm Mattis}} \int dm \tanh(\beta \xi_i m) \tanh(\beta \xi_j m) \exp\left( -\beta \Nv\frac{m^2}{2}+\sum_k \log 2 \cosh(\beta \xi_k m)\right)
\end{align*}
Evaluating the saddle point of the argument of the exponential (which is the same as the one for the partition function) we have that
\begin{equation*}
    \langle v_i v_j \rangle_{\mathcal{D}} = \xi_i \xi_j m^2 \text{ where } m = \tanh\left(\beta m \right)
\end{equation*}
The negative term can also be computed in the thermodynamic limit
\begin{align*}
    \langle v_i v_j \rangle_{\rm RBM} &= \frac{1}{Z_{\rm RBM}} \int dh\sum_{\bm{v}} v_i v_j \exp\left( \sum_k v_k h w_k - \frac{h^2 \Nv}{2} \right) \\
    &=\int dh \frac{1}{Z_{\rm RBM}} \tanh(h w_i) \tanh(h w_j) \exp\left( \sum_k \log\left[2 \cosh(h w_k) \right] - \frac{h^2 \Nv}{2} \right) \\
    &= \tanh(h^* w_i) \tanh(h^* w_j) \text{ with } h^* = \frac{1}{\Nv}\sum_k w_k \tanh(h^* w_k) 
\end{align*}
where the last line is obtain by taking the saddle point of the integral over $h$, ($h^*$ corresponding to the extremum). We can now express the gradient as
\begin{align*} 
    \frac{dw_i}{dt} &= \frac{1}{\Nv} \xi_i  \sum_k \xi_k w_k  m^2 - \frac{1}{\Nv}\sum_k w_k \tanh(h^* w_k) \tanh(h^* w_i) \\
                    &= \frac{1}{\Nv} \xi_i  \sum_k \xi_k w_k  m^2 - h^* \tanh(h^* w_i) 
\end{align*}
Assuming first that the weights are small we get that $h^* = 0$. We can solve the gradient's equations in this regime. In such case, the only solution for the saddle point equation of the RBM is given by $h^* = 0$ and we can see that the solution of the evolution of the weight is global toward the direction $\bm{\xi}$ by projecting the differential equation on the preferred direction. Defining $U_{\bm{\xi}} = \Nv^{-1/2}\sum_i \xi_i w_i$, we obtain
\begin{equation*}
    \frac{dU_{\bm{\xi}}}{dt} = m^2 U_{\bm{\xi}} \text{ thus } U_{\bm{\xi}} = U_{\bm{\xi}}^0 e^{m^2 t}.
\end{equation*}
This shows that the weights are growing in the direction of $\bm{\xi}$ while the projection on any orthogonal direction $\bm{\phi}^\alpha$ stays constant: 

\begin{equation*}
    \bm{\phi}^\alpha \cdot \frac{d\bm{w}}{dt} = \frac{m^2}{\Nv} (\bm{\phi}^\alpha \cdot \bm{\xi}) (\bm{w} \cdot \bm{\xi}) = 0 \text{ since } \bm{\phi}^\alpha \cdot \bm{\xi} = 0
\end{equation*}
When the weights grow larger, the solution for $h^*$ will depart from zero. The correlation of the learning RBM then starts to grow
\begin{align*}
    \langle v_i v_j \rangle_{\rm RBM} &\approx \frac{1}{Z} \int dh h^2 w_i w_j \exp\left( -\frac{\Nv h^2}{2} + \sum_k \frac{h^2 w_k^2}{2}\right) = w_i w_j \frac{1}{\Nv\left(1-\sum_k w_k^2 / \Nv  \right)} \\
    \chi &= \sum_{i,j} \xi_j \xi_i \langle v_i v_j \rangle_{\rm RBM} \approx \left( \sum_i \xi_i w_i \right)^2 \frac{1}{\Nv\left(1-\sum_i w_i^2 / \Nv  \right)}
\end{align*}
and diverges when $\Nv^{-1}\sum_i w_i^2 \sim 1$, therefore exhibiting a second order phase transition during the learning. Finally, we can study the regime where the weights are not small. In that case, we can first observe that the evolution of the directions orthogonal to $\bm{\xi}$, $\bm{\phi}^\alpha$ are given by
\begin{equation*}
    \sum_i \phi_i^{\alpha} \frac{dw_i}{dt} = \frac{m^2}{\Nv} \sum_i \phi_i^\alpha \xi_i  \sum_{k} \xi_k w_k  - \sum_i \phi_i^\alpha  h^* \tanh(h^* w_i) = - \sum_i \phi_i^\alpha  h^* \tanh(h^* w_i)
\end{equation*}
which will cancel if the weight $\bm{W}$ aligns totally with the $\bm{\xi}$. Finally, taking $w_i = \xi_i w, $ the gradient projected along $\bm{\xi}$ at stationarity imposes
\begin{equation*}
    w m^2 = h^* \tanh(h^* w) \text{ and thus } w=\sqrt{\beta} \text{ and } h^* = \sqrt{\beta m}
\end{equation*}

\section{Binary-Binary RBM}\label{SI:BinaryBinaryRBM}
\begin{figure}
    \centering
    \includegraphics[scale=0.47]{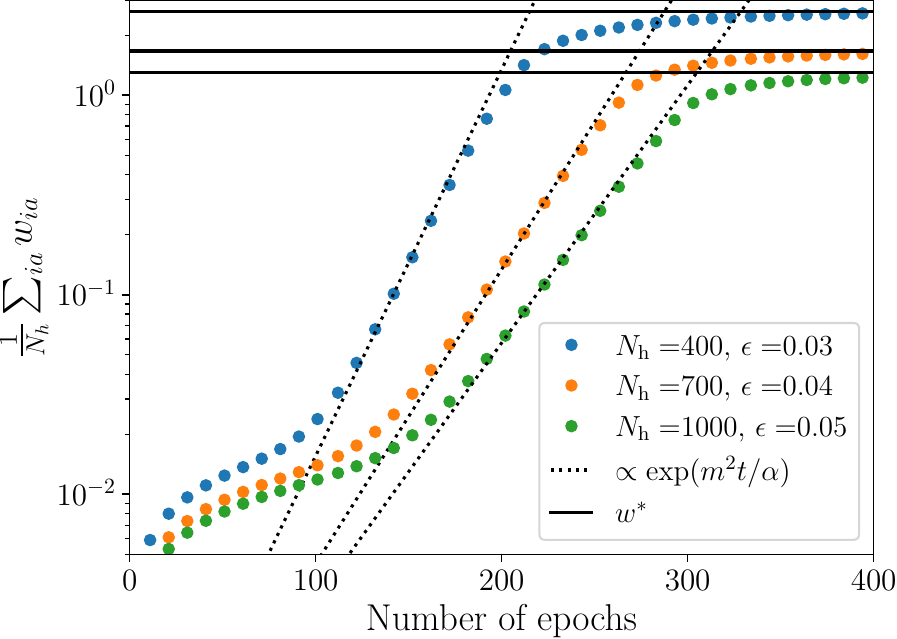}
    \includegraphics[scale=0.47]{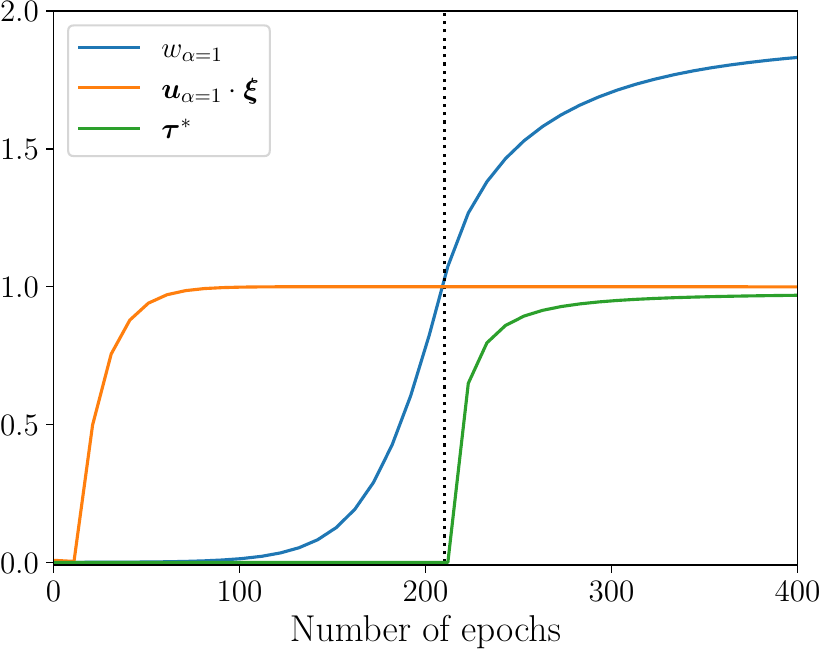}
    \caption{\textbf{Left:} learning behavior of the Binary-Binary RBM, using data from the Mattis model. The different curves correspond to systems of size $\Nv=900$ at inverse temperature $\beta=1.4$ with learning rate $\epsilon=0.03,0.04,0.05$ and $\Nh =400,700,1000$ respectively. The argument of the exponential curves are not adjusted but set to $m^2 \epsilon / \alpha$. \textbf{Right:} we illustrate the RBM's dynamics in the binary-binary case with $\beta=1.4$ and $\Nv=900$, $\Nh=400$. First the eigenvector $\bm{u}^{\alpha=1}$ aligns itself with the pattern $\bm{\xi}$. Then, the eigenvalue $w_{\alpha=1}$ grows exponentially until reaching saturation and when it crosses the value $1$, the system develops a spontaneous magnetization.}
    \label{fig:DynBinBin}
\end{figure}

The RBM sharing both discrete binary variables on the visible and hidden nodes is by far the most commonly used. In particular, using binary nodes in the hidden layer instead of the Gaussian distribution allows the model to potentially fit any order correlations of the dataset. In this section, we review how the learning dynamics translate to this case, using for simplicity binary $\{\pm 1\}$ variables. In order to obtain an interesting behavior in this phase of the learning, it is important to consider a particular parametrization of the RBM. We consider that all hidden nodes share the same weight. This is important to be able to have a recall phase transition in the model. We therefore have the following Hamiltonian
\begin{equation}
    \mathcal{H} = -\frac{1}{\Nh} \sum_i v_i w_i \sum_a h_a
\end{equation}
where $\Nh=\alpha \Nv$ is the number of hidden nodes $h_a$ of the system and the vector $\bm{W}$ correspond to the weight shared across all the hidden nodes. In this model, we can now compute the positive and negative of the gradient. The first one is given by
\begin{align*}
    \frac{1}{N_{h}}\langle v_{i}&\sum_{a}h_{a}\rangle_{\mathcal{D}} = \frac{1}{Z_{\rm Mattis}} \sum_{\bm{v}} \int dm v_i\exp\left( -\frac{\beta m^2 \Nv}{2} + m \beta \sum_j \xi_j v_j \right) \frac{1}{\Nh}\sum_a \tanh\left[\Nh^{-1} \sum_j w_j v_j \right] \\
    &=  \frac{1}{Z_{\rm Mattis}} \sum_{\bm{v}} \int dm d\tau v_i \exp\left( -\frac{\beta m^2 \Nv}{2} + m \beta \sum_j \xi_j  v_j \right) \delta(\tau - \Nh^{-1} \sum_j w_j v_j) \tanh(\tau) \\
    &=  \frac{1}{Z_{\rm Mattis}} \sum_{\bm{v}} \int d\tau d\bar{\tau} dm v_i \exp\left( -\frac{\beta m^2 \Nv}{2} + m \beta \sum_j \xi_j v_j \right) \tanh(\tau) e^{i\tau \bar{\tau} - i\Nh^{-1 }\bar{\tau}\sum_j w_j v_j} \\
    &= \frac{1}{Z_{\rm Mattis}}  \int dm d\tau d\bar{\tau} e^{-\beta m^2 \Nv/2+i\tau \bar{\tau}} \tanh(\xi_i m\beta -i\Nh^{-1 }\bar{\tau}w_i) \tanh(\tau) \\
    & \;\;\;\;\;\;\;\;\;\;\;\; \times\exp\left(\sum_j \log \cosh\left[\xi_j\beta m -i\Nh^{-1 }\bar{\tau} w_j \right] \right) 
\end{align*}
finding the saddle point of the argument in the exponential, we obtain
\begin{align*}
    \frac{1}{N_{h}}\langle v_{i}\sum_{a}h_{a}\rangle_{\mathcal{D}} = \xi_i \tanh(\beta m) \tanh\left(\frac{m}{\Nh} \sum_j \xi_j w_j\right) = \xi_i m\tanh\left(\frac{m}{\Nh} \sum_j \xi_j w_j\right)
\end{align*}
The same type of computation can be done for the negative term, we found that
\begin{align*}
\frac{1}{N_{h}}\langle v_{i}\sum_{a}h_{a}\rangle_{RBM} & =\xi_i \tau\tanh\left(w_{i}\tau\right) \\
\tau & = \tanh\left(\frac{1}{N_{h}}\sum_{j}w_{j}\tanh\left(w_{j}\tau\right)\right)
\end{align*}
Again, in the small coupling regime (or at the beginning of the learning), when $\Nh^{-1} \sum_j w_j^2 \ll 1$, we have that $\tau = 0$. In such case, the gradient over the weight matrix is given by
\begin{equation*}
    \frac{dw_i}{dt} = \xi_i m \tanh\left( \frac{m}{\Nh} \sum_j \xi_j w_j \right)
\end{equation*}
following the same approach as in the main text, we project the weights on the unit vector $\bm{u}_1 = \bm{\xi}/\sqrt{\Nv}$, $U_{\bm{\xi}} = \bm{u}_1 \bm{W}$, which gives
\begin{equation*}
    \frac{dU_{\bm{\xi}}}{dt} = \sqrt{\Nv} m \tanh\left( \frac{m\sqrt{\Nv}}{\Nh}U_{\bm{\xi}} \right)
\end{equation*}
We can integrate this equation, obtaining the solution
\begin{align*}
    \sinh\left( \frac{m}{\sqrt{\Nv}\alpha}U_{\bm{\xi}}(t) \right) &= \sinh\left( \frac{m}{\sqrt{\Nv}\alpha}U_{\bm{\xi}}(0) \right) \exp\left(\frac{m^2 t}{\alpha} \right) \\
    U_{\bm{\xi}}(t) &= U_{\bm{\xi}}(0) \exp\left(\frac{m^2 t}{\alpha} \right) 
\end{align*}
where the second line is obtained in the very large $\Nv$ limit. Again we have an exponential growth in the first steps of the learning. At the end of the learning, the weights again align in the direction of $\bm{\xi}$. This can be checked by the fact that the positive term of always orthogonal to any vector orthogonal to $\bm{\xi}$, and thus the simplest option for the gradient projected in those direction is to be orthogonal to $\bm{\xi}$. Taking $\bm{W} = \bm{u}_1 w$, we obtain
\begin{align*}
    m\tanh\left(mw/\alpha\right) &= \tau  \tanh(w \tau) \\
    \tau &= \tanh(w\tanh(w\tau)/\alpha) \\
\end{align*}
The solution can be found numerically by solving the fixed point equation on $\tau$, and measuring the magnetization of the dataset. In Fig.\ref{fig:DynBinBin} we illustrate our results in the same dataset as in the section~\ref{sec:dynamicalGauss}, taking the Mattis model with $\Nv=900$, $\beta=1.4$, varying the learning rate and the number of hidden nodes.

\section{Learning with correlated patterns} \label{SI:corr_patt}

In this part we detail how the learning goes when considering a pair of correlated patterns. As described in \ref{sec:corr_patt}, the pairs of patterns are defined as 
\begin{equation*}
    \bm{\xi}^1 = \bm{\eta}^1 + \bm{\eta}^2 \text{ and } \bm{\xi}^2 = \bm{\eta}^1 - \bm{\eta}^2
\end{equation*}
where $\bm{\eta}^1$ is a vector whose first $\Nv \frac{1+\kappa}{2}$ components are equal to $\pm 1$ with equal probability and the remaining ones are zero. The other vector $\bm{\eta}^2$ has its last $\Nv \frac{1-\kappa}{2}$ components equal to $\pm 1$ with equal probability and the rest are 0; we also have that $\kappa \in [0,1]$. When $\kappa=1$, both patterns $\bm{\xi}^{1,2}$ are equal, while otherwise different but correlated. In particular $\mathbb{E}_{\bm{\eta}^1,\bm{\eta}^2}[\bm{\xi}^1 \bm{\xi}^2 ] = \Nv\kappa$. Following the results of~\cite{tamarit1991pair}, it is possible to compute the saddle point equations for the magnetization. The general form is given by
\begin{align*}
    m_1 &= \frac{1}{\Nv} \sum_i \xi^1_i \tanh\left( \beta m_1 \xi^1_i + \beta m_2 \xi^2_i\right) \\
    m_2 &= \frac{1}{\Nv} \sum_i \xi^2_i \tanh\left( \beta m_1 \xi^1_i + \beta m_2 \xi^2_i\right) \\
\end{align*}
This system has been solved in~\cite{tamarit1991pair} and exhibits the following properties. When $T>1+\kappa$, the system is in the paramagnetic regime and $m_1=m_2=0$. When the temperature is lowered and lies in $1-\kappa<T<1+\kappa$, the solution is given by the pair retrieval $m_1=m_2=m=\frac{1+\kappa}{2}\tanh(2\beta m)$. Finally, when $T<1-\kappa$, the system condensates on the following solution
\begin{align*}
    m_1 &= \frac{1+\kappa}{2}\tanh\left( \beta(m_1+m_2)\right) + \frac{1-\kappa}{2}\tanh\left( \beta(m_1-m_2)\right) \\
    m_2 &= \frac{1+\kappa}{2}\tanh\left( \beta(m_1+m_2)\right) - \frac{1-\kappa}{2}\tanh\left( \beta(m_1-m_2)\right)
\end{align*}
where basically the system either condensates toward one of the pattern $\bm{\xi}^{1,2}$, while the other magnetization has some non-zero value due to the correlation. 

\begin{figure}
    \centering
    \includegraphics[scale=0.47]{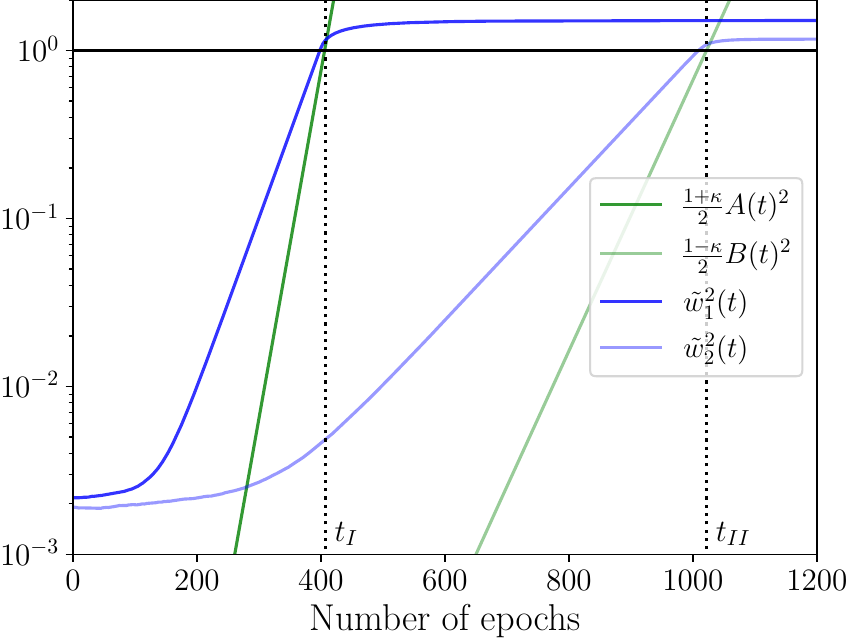}
    \includegraphics[scale=0.47]{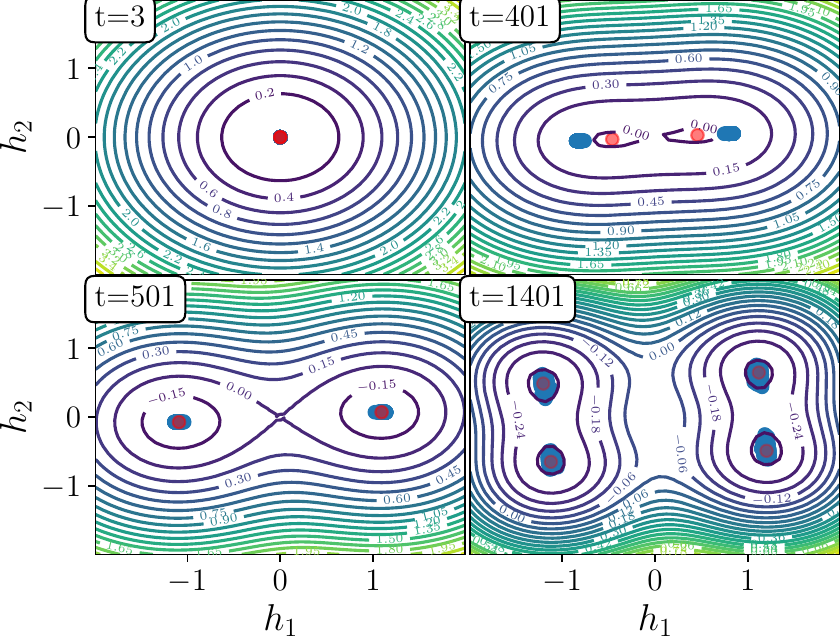}
    \caption{\textbf{Left:} the empirical dynamics of the eigenvalues of the weight matrix, here denotes $\tilde{w}_\alpha$ in blue. In green, the predicted dynamics as in eq.~\ref{eq:dyn_weigth}, adjusting only the initial conditions $U_{\bm{\eta}^1}^{a}(0)$ and $U_{\bm{\eta}^2}^{a}(0)$. We see that the curves cross the line $y=1$ at the same moments $t_I$ and $t_{II}$. \textbf{Right:} the free energy in the plane $(h_1,h_2)$, the order parameters of the model. For different value of the weights during learning, we reconstruct the free energy of the system. We clearly see how the RBM first creates two minima, in the direction of $\bm{\eta}^1$, and then, split again to obtain the four fixed points.}
    \label{fig:dyn_PT}
\end{figure}

We can use the thermodynamics properties of this model to study how the learning of the RBM should behave in the regime $T<1-\kappa$. In order to use this model as generating the dataset, we need to compute the correlations $\langle s_i s_j \rangle$. The model presents four fixed points, all equally probable:
\begin{align*}
    (m_1,m_2) &= (m^+,m^-) \text{ and its symmetric case } (m_1,m_2) = (-m^+,-m^-) \\
    (m_1,m_2) &= (m^-,m^+) \text{ and its symmetric case } (m_1,m_2) = (-m^-,-m^+) 
\end{align*}
where $m^+>m^- >0$. Therefore, writing $r=\tanh(\beta(m^+ + m^-))$ and $p=\tanh(\beta(m^+ - m^-))$ we have that
\begin{align*}
    \langle v_i v_j\rangle_{data} &= \frac{1}{4}\left(\sum_{(m_1,m_2)} \left[(\eta_i^1+\eta_i^2) \tanh(\beta(m_1+m_2))\right]\left[(\eta_j^1+\eta_j^2)  \tanh(\beta(m_1+m_2)) \right]\right) \\
    &= \eta_i^1 \eta_j^1 r^2 + \eta_i^2 \eta_j^2 p^2
\end{align*}
because the cross terms $\eta_i^1 \eta_j^2$ are canceled when changing $(m_1=m^+,m_2=m^-)$ to $(m_1=m^-,m_2=m^+)$.
At this point, it is possible to write the gradient at the linear order and project it toward both direction $\bm{\eta}^1$ and $\bm{\eta}^2$. Denoting $U_{\bm{\eta}^1}^{a} = \bm{\eta}^1\cdot \bm{w}^{a}$ and $U_{\bm{\eta}^2}^{a} = \bm{\eta}^2\cdot \bm{w}^{a}$, we get
\begin{align*}
    \frac{dU_{\bm{\eta}^1}^{a}}{dt} &= r^2\frac{1+\kappa}{2} U_{\bm{\eta}^1}^{a} \\
    \frac{dU_{\bm{\eta}^2}^{a}}{dt} &= p^2\frac{1-\kappa}{2} U_{\bm{\eta}^2}^{a}
\end{align*}
Using this form, we end up with the following solution of the weight matrix
\begin{equation}
    w_i^{a}(t) = w_i^{a}(0) + \frac{\eta_i^1 U_{\bm{\eta}^1}^{a}(0)}{(1+\kappa)/2}\left[\exp\left(r^2\frac{1+\kappa}{2}t\right) - 1\right] + \frac{\eta_i^2 U_{\bm{\eta}^2}^{a}(0)}{(1-\kappa)/2}\left[\exp\left(p^2\frac{1-\kappa}{2}t\right) - 1\right] \label{eq:dyn_weigth}
\end{equation}
We therefore understand the following. At the beginning of the learning, since $r>p$, what is learned first is the mode toward the direction $\bm{\eta}^1 \propto \bm{\xi}^1+\bm{\xi}^2$, in a timescale that is given by time $t \sim 1/r^2$. At a different timescale, the part that is aligned with $\bm{\eta}^2$ will grow as well as discussed in the main text. 
Following the dynamics of the weights, as in eq.~\ref{eq:dyn_weigth}, we can infer the moment where the phase transition occurs. When considering an Hopfield like model, we know that the transition happens when
\begin{equation*}
    \frac{\beta}{2 \Nv} \left( \sum_i v_i \xi_i \right)^2 \sim \frac{1}{2 \Nv} \left( \sum_i v_i \xi_i \right)^2
\end{equation*}
that is, the critical temperature is $\beta_c=1$. Following the dynamics of the weights of eq.~\ref{eq:dyn_weigth}, and neglecting the terms that are not aligned with $\bm{\eta}^1$ we can write
\begin{align*}
    \frac{1}{2 \Nv} \left( \sum_i v_i \xi_i \right)^2 &\sim \left(\frac{ U_{\bm{\eta}^1}^{a}(0)}{(1+\kappa)/2}\right)^2\left[\exp\left(r^2\frac{1+\kappa}{2}t\right) - 1\right]^2 \frac{1}{2 \Nv} \left( \sum_i v_i \eta_i^1 \right)^2 \\
    \beta_{\bm{\eta}^1}(t) &= \left(\frac{ U_{\bm{\eta}^1}^{a}(0)}{(1+\kappa)/2}\right)^2\left[\exp\left(r^2\frac{1+\kappa}{2}t\right) - 1\right]^2
\end{align*}
where we can identify a sort of dynamical temperature associated to the pattern $\bm{\eta}^1$. Now, we need to be careful since by definition, the pattern $\bm{\eta}^1$ is made of random $\pm 1$ components on its $(1+\kappa)/2$ elements and zero elsewhere. This rescales the critical temperature by a factor $(1+\kappa)/2$. Therefore we need to look when
\begin{equation*}
    \frac{1+\kappa}{2}\beta_{\bm{\eta}^1}(t_I) \sim 1
\end{equation*}
and the same kind of argument can be used for the second transition with this time
\begin{align*}
    \beta_{\bm{\eta}^2}(t) &= \left(\frac{ U_{\bm{\eta}^2}^{a}(0)}{(1-\kappa)/2}\right)^2\left[\exp\left(p^2\frac{1-\kappa}{2}t\right) - 1\right]^2 \\
    \frac{1-\kappa}{2}\beta_{\bm{\eta}^2}(t_{II}) &\sim 1
\end{align*}
We show in Fig.~\ref{fig:dyn_PT}, left panel how the times $t_I$ and $t_{II}$ compare with the moment where the eigenvalues $w_\alpha$ of the weight matrix cross the value one, which correspond to the phase transition following a statistical mechanics approach~\cite{decelle2018thermodynamics}. We observe that both indicators are crossing the line $y=1$ at the same moment. In Fig.~\ref{fig:dyn_PT}, right panel, we plot the behavior of the free energy (in the plane $(h_1,h_2)$). We see that at the moment of the transition, the free energy opens in the direction corresponding to the transition. Projecting the dataset (black dots) in the same direction as $h1$ (resp. $h2$), we can see how the system correctly positioned the minima once fully trained.

\section{The datasets and the rescaling} \label{sec:datasets}
In this work, we illustrated our results on three datasets:
\begin{enumerate}
    \item The Human Genome Dataset (HGD) ~\cite{10002015global} containing binary vectors, each representing a selection of $805$ genes from a human individual, where 1s or 0s indicate the presence or absence of gene mutations relative to a reference sequence.
 \item The MNIST dataset~\cite{lecun1998gradient}, containing $ 28\times 28$ pixel black and white images of digitized handwritten digits.
 \item The CelebA~\cite{DBLP:journals/corr/abs-1710-10196} dataset, in black-and-white, with $128\!\times\! 128$ pixel images of celebrities faces.
\end{enumerate}

The datasets MNIST and CELEBA, were either downscaled or upscaled in order to create dataset of various sizes. In practice, the function \textit{resize} from the python library skimage was used either to increase or decrease the image size. The dataset HGD is geometrically a one-dimensional structure. In order to reduce its size, we took the convolution of each sample with a kernel of size $s=3$. The output is one if the sum of the three input values (that are discrete variables in $\{0,1\}$) of the kernel is above the threshold 2 and zero otherwise. A stride of $2$ has been chosen such that the resulting samples has its size reduced by a factor two.

\section{Details on the training and the numerical analysis}

\subsection{Training}\label{sec:training}
All RBMs analyzed in the main text were trained with the Persistent Contrastive Divergence (PCD) method~\cite{tieleman2008training} and $k\!=\!100$ MCMC sampling steps per parameter update,
 to approximate the negative term of the log-likelihood gradient in Eq.~\eqref{eqs:grad3}. In this scheme, the last configurations reached in the MCMC process to compute the previous update are used as initialization of the new chains used to compute the subsequent update. This scheme tends to favor the equilibrium regime~\cite{decelle2021equilibrium}. The results with other training schemes is discussed in section~\ref{app:noneq}.
 Moreover, as usual, we keep $N_\mathrm{chains}$ independent parallel Markov chains. For simplicity, $N_\mathrm{chains}$ is chosen to match the minibatch size used to estimate the positive term of the gradient. The code to reproduce the experiments is freely available in \href{GitHub:AurelienDecelle/TorchRBM}{https://github.com/AurelienDecelle/TorchRBM}. The hyperparameters used for each training (no. of visible and hidden units $\Nv$ and $\Nh$, respectively,  learning rate $\epsilon$ or minibatch size $N_{\rm ms}$) are given in Table~\ref{tab:trainings}.
 \begin{table}[t!]
     \centering
     \begin{tabular}{ccc|cc}
          dataset&$\Nv$&$\Nh$& $\epsilon$ &$N_{\rm ms}$\\\hline
          MNIST & $14\times 14$& 250& 0.005 &500\\
          MNIST & $19\times 19$& 500& 0.005 &500\\
          MNIST & $28\times 28$& 1000& 0.005 &500\\
          MNIST & $39\times 39$& 2000& 0.005 &500\\
          MNIST & $56\times 56$& 4000& 0.001 &500\\\hline
          CelebA & $45\times 45$& 125& $4\cdot 10^{-5}$ &500\\
          CelebA & $64\times 64$& 250& $4\cdot 10^{-5}$ &500\\
          CelebA & $90\times 90$& 500& $4\cdot 10^{-5}$  &500\\
          CelebA & $128\times 128$& 1000& $4\cdot 10^{-5}$ &500\\\hline
          HGD & 201& 25& 0.002 &4500\\
          HGD & 402& 50& 0.002 &4500\\
          HGD & 805& 100& 0.002 &4500\\
     \end{tabular}
     \caption{Hyperparameters of the RBMs analyzed in the main-text.}
     \label{tab:trainings}
 \end{table}
 \subsection{Numerical analysis}\label{sec:num-analysis-details}
 \subsubsection{Susceptibility}
 Part of the analysis in section~\ref{sec:num_ana} of the main text is based on sampling the equilibrium configurations of models trained at different epochs to extract the moments of the distribution of magnetizations $m_\alpha$, i.e. the projection of the samples along the different $\alpha$-th singular vectors of $\bm W$. For this purpose, we automatically selected $10^3$ models uniformly in logarithmic scale in training time and annealed the $N_\mathrm{s}=1000$ independent samples from the least trained model to the most trained model, following the hot-to-cold thermal analogy, i.e. we perform $N_\mathrm{mesfr}$ alternate Gibbs sampling MCMC steps on each set of model parameters  and use the last achieved configurations as a starting point to initialize the run on the next set of model parameters. The less trained model is initialized randomly. In parallel, we also consider the reverse scheme, where we consider a heating annealing. We start with the most trained machine, where all visible units are initialized to 1 (to force all initial configurations to be in a single cluster), and move backwards it in training time. We systematically checked that both analyses gave the same results for $N_\mathrm{mesfr}=1000$ in the region of interest (the critical region) in Fig.~\ref{fig:MNIST} and~\ref{fig:CELEBA}.

In addition to the sampling procedure, we computed the SWD of the matrix $\bm W$ at each new parameter set and projected each visible configuration $\bm v$ along each of the left singular vectors $\bm u_{\alpha}$, to obtain $m_\alpha=(\bm u_{\alpha}\cdot \bm v)/\sqrt{\Nv}$, and each hidden configuration $\bm h$ along the right singular $\bar {\bm u}_{\alpha}$ vector, to obtain $\bar m_\alpha=( \bar{\bm u}_{\alpha}\cdot \bm v)/\sqrt{\Nv}$. The distribution moments are later estimated using the sample mean and variance of these $N_\mathrm{s}$ measures.  

\subsubsection{Hysteresis loop}\label{app:histnum}
To investigate the hysteresis behavior between different lumps at or above the phase transition, we first select the training epochs in which the first singular values of $\bm W$, $w_1$, take the values 4, 4.4 and 5. Note that for MNIST $w_\mathrm{1,c}$ is approximately $4$. With the model parameters at each of these three selected number of updates, we perform $N_\mathrm{s}$ independent MCMC runs with the tilted Hamiltonian from Eq.~\eqref{eq:hyst}. In these runs, the external field $h$ is gradually varied to trace a loop: starting from $h=0$, we slowly increase it to $h_\mathrm{max}$, then decrease it to $-h_\mathrm{max}$ and finally bring it back to $h=0$. Again, the last configurations reached at a certain $h$ are used as initialization for the next one. In practice, we have chosen $h_\mathrm{max}/\Nv^{0.75}$, the increment in the field as $\delta h=2\times h_\mathrm{max}/N_{\rm loop}$ and $N_{\rm loop}=50$. We can modulate the speed of the loop by performing a different number of MCMC steps $k$ at each value of $h$. As shown in the figure, we consider $k=10$, 100, 1000 and $10^4$.
The results shown in the main text were obtained using the RBM trained with the original CELEBA dataset (i.e. $\Nv=128\times 128$), but the results obtained with other sizes and datasets are completely analogous.

\section{Link between the low-rank RBM and Mattis model} \label{app:RBM_Mattis}

Let us consider a low-rank Ising-Ising RBM in which the $\mathcal{\bm W}$ matrix has a single non-zero singular value $\omega$, with left and right singular vectors $\bm u$ and $\bar{\bm {u}}$, and visible and hidden Ising variables (let's call these variables $\mathcal{W}$, $\bm s$ and $\bm \tau$ to distinguish them from the binary ${0,1}$ version, which would be $\bm{W}$ and $\bm v$ and $\bm h$). In this case, the energy function of the RBM (if we ignore the biases for now) is 
\begin{equation*}
    E(\bm v,\bm h)=- \omega (\bm u\cdot \bm v)(\bar{\bm u}\cdot \bm h),
\end{equation*}
which leads to a marginal energy on the visible
\begin{eqnarray}
    \mathcal{E}(\bm s)=-\sum_a \log \cosh \caja{\sqrt{\Nv}w \bar{u}_a^2 m} \approx -\frac{1}{2}\Nv\omega^2 m^2+O(m^4),
\end{eqnarray}
where we have defined $m=\bm u\cdot \bm v/\sqrt{\Nv}$ as the magnetization of the spins along the direction $\bm u$ and have exploited the fact that $\sum_a \bar{u}_a^2=1$ because it is a unit vector. One can obtain an analogous expression for the marginal energy on the hidden units, formulated in terms of the hidden magnetization $\bar m=\bar{\bm u}\cdot \bm \tau/\sqrt{\Nh}$.
These energy functions, for small $m$ or $\bar m$, are formally equal to those of the Mattis model for $\beta=\omega^2$, which means that our RBM should manifest a critical phase transition at $\beta_\mathrm{c}=T_\mathrm{c}^{-1}=\omega_\mathrm{c}^2=1$, with mean-field critical exponents. Standard RBMs are not formulated as Ising $\pm 1$ variables, but in the form of binary $\{0,1\}$ variables where we have the equivalence $4 \bm{\mathcal{W}}=\bm{W}$ between the couplings matrices. This results in a critical point at $w_\mathrm{c}=4$ and an effective inverse temperature $\beta=w^2/16$.

\section{Hysteresis in discontinuous transitions} \label{SI:hyst}

The hysteresis phenomenon shown in Fig.~\ref{fig:CELEBA}-E is a classical measure in statistical physics and is a unique signature developed and observed in statistical physics to reveal that a high-dimensional probability measure had a phase transition where it splits into two distinct lumps. The procedure is to tilt the probability measure by introducing a contribution in the energy function that favors one lump over the other. In the present case, we use the learned preferred direction associated with the first phase transition $\bm{u}$, which is also the direction in which the probability measure splits, and we add a magnetic field to break the symmetry between the two modes created by the learning process. Therefore, we tilt the measure by adding
\begin{equation*}\label{eq:hyst}
    \mathcal{H}^{\rm tilted} = \mathcal{H}^{\rm RBM} - h \sum_i u_i v_i.
\end{equation*}
The tilted Hamiltonian favors one of the two lumps depending on the value of the local bias $h$. When the measure is actually concentrated on two distinct lumps, one of them leads to a sudden discontinuous transition (``first-order transition” in physics). In our case, the lumps are associated with the learned patterns and this additional contribution consists of the scalar product between the visible variables and the times of the learned patterns (the field that controls the strength of the tilting). In the presence of a first-order phase transition, one usually finds the phenomenon of hysteresis, i.e. the transition from one clump to another can be delayed due to metastability, leading to the characteristic hysteresis loops that we show in Fig.~\ref{fig:CELEBA}-E (see e.g.~\cite{den2004metastability,Bovier2009} for a rigorous treatment and~\cite{chaikin1995principles} for a physical treatment). This figure thus provides direct evidence for the decomposition of the measure into distinct lumps corresponding to the learned patterns, and that this decomposition occurs at the second-order phase transition that occurs during learning. The details about the numerical implementation are given in Section~\ref{app:histnum}.

\section{Results with other training schemes}\label{app:noneq}
All the machines analyzed in the main text were trained using the PCD-100 scheme, which involves initializing the chains with the PCD method and performing 100 MCMC steps per parameter update. This approach ensures that we obtain models in good equilibrium, avoiding the non-monotonic behavior in sample quality typical of out-of-equilibrium regimes~\cite{decelle2021equilibrium}. However, it is more computationally expensive than standard methods, where only $k=1-10$ steps are used, or alternative initialization strategies like Contrastive Divergence (CD)~\cite{hinton2002training}, where chains are initialized using the minibatch samples at each update, or the fully out-of-equilibrium regime (Rdm), where chains are always initialized randomly~\cite{decelle2021equilibrium, nijkamp2019learning}.

\begin{figure}[!t]
    \centering
    \includegraphics[height=8cm]{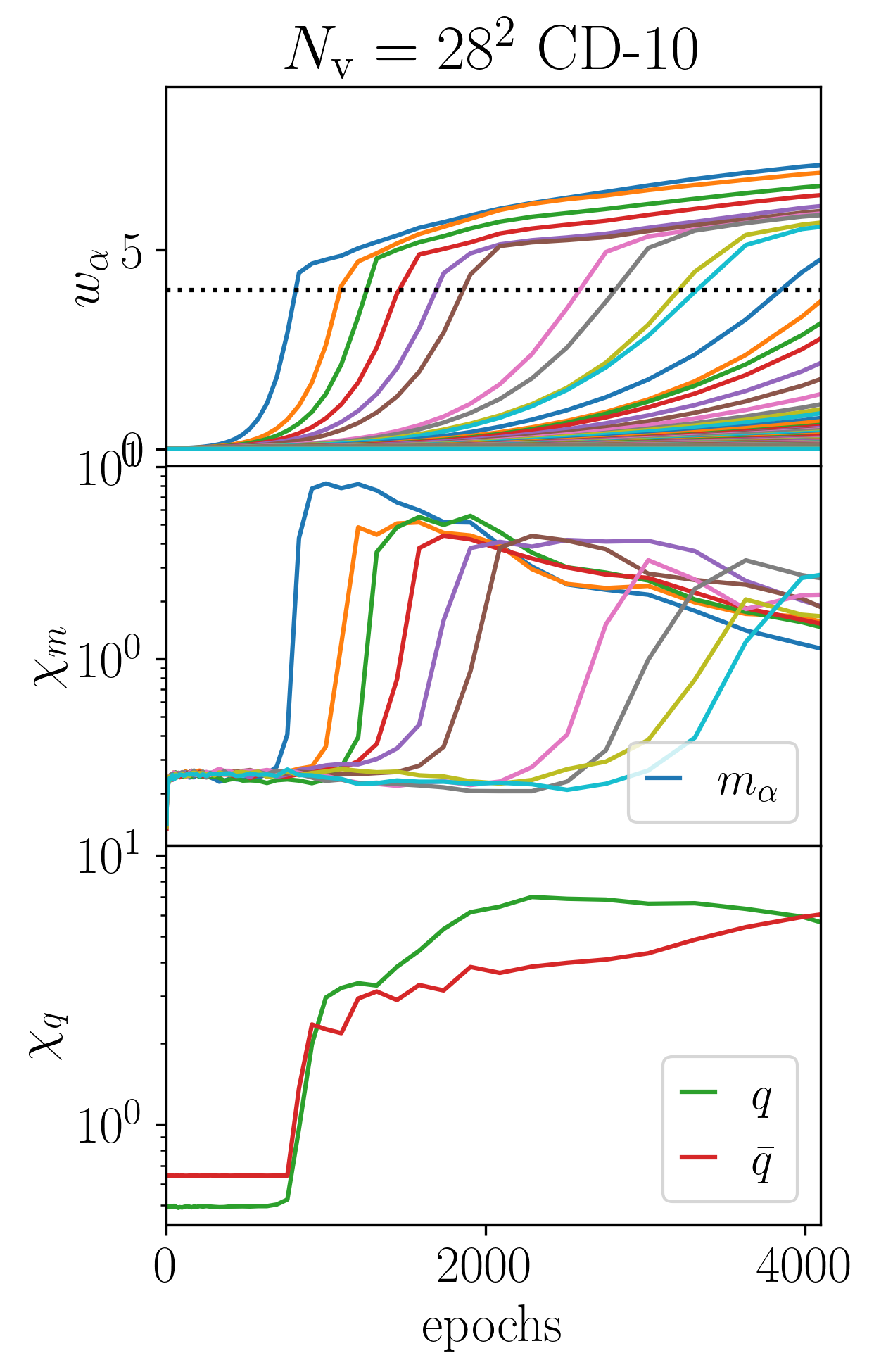}
    \includegraphics[height=8cm]{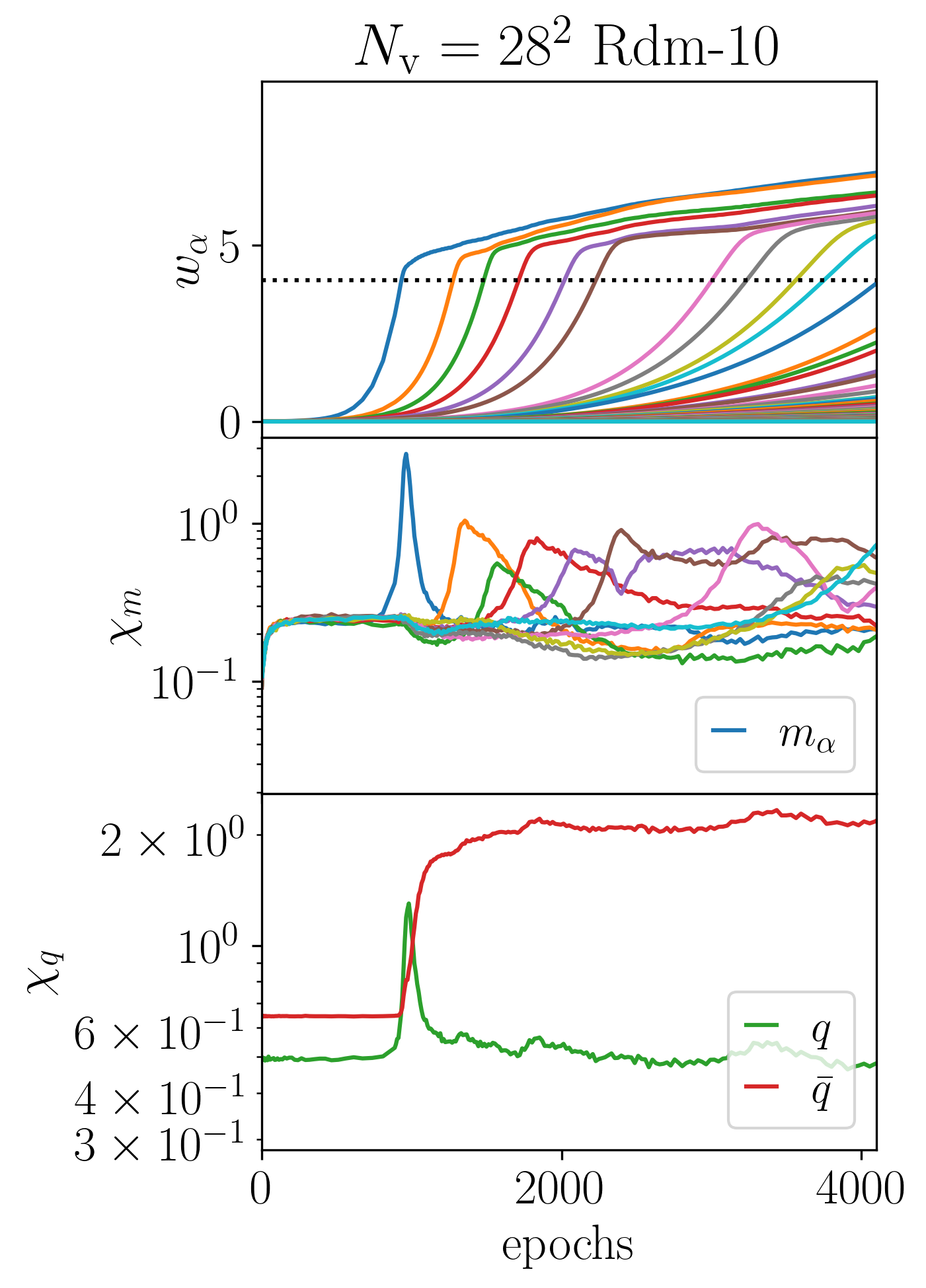}
    \caption{We reproduce Figs.~\ref{fig:MNIST}A,B and C but for RBMs trained with CD-10 (left) and Rdm-10 (right).
    }
    \label{fig:ooe}
\end{figure}

In this appendix, we analyze RBMs trained with CD-10 and Rdm-10 strategies on the MNIST dataset. While the time evolution differs -often degrading susceptibility along learning directions— the overall picture of the cascade of phase transitions remains unchanged. We show the equivalent to Figs.~\ref{fig:MNIST}A,B,C with these two new trainings in Fig.~\ref{fig:ooe}.

\newpage

\end{document}